\documentclass[10pt,twocolumn,letterpaper]{article}

\usepackage[pagenumbers]{cvpr} % To force page numbers, e.g. for an arXiv version

\usepackage[accsupp]{axessibility} % Improves PDF readability for those with disabilities.

\usepackage{graphicx}
\usepackage{amsmath}
\usepackage{amssymb}
\usepackage{booktabs}

\usepackage{enumitem}
\usepackage{pifont}% http://ctan.org/pkg/pifont
\newcommand{\cmark}{\ding{51}}%
\newcommand{\xmark}{\ding{55}}%
\usepackage{amsthm}

\usepackage{multirow}
\usepackage[table, svgnames, dvipsnames]{xcolor}
\definecolor{LightCyan}{rgb}{0.88,1,1}

\usepackage{rotating}
\usepackage{gensymb}

\usepackage[pagebackref,breaklinks,colorlinks]{hyperref}

\usepackage[capitalize]{cleveref}
\crefname{section}{Sec.}{Secs.}
\Crefname{section}{Section}{Sections}
\Crefname{table}{Table}{Tables}
\crefname{table}{Tab.}{Tabs.}

\def\cvprPaperID{1413} % *** Enter the CVPR Paper ID here
\def\confName{CVPR}
\def\confYear{2023}

\begin{document}

\title{\vspace{-20pt}
3D Semantic Segmentation in the Wild: 

Learning Generalized Models for Adverse-Condition Point Clouds\vspace{-15pt}
}

% \maketitle
\author{Aoran Xiao$^{1}$, Jiaxing Huang$^{1}$, Weihao Xuan$^{2}$, Ruijie Ren$^{3}$, Kangcheng Liu$^{1}$\\
Dayan Guan$^{4}$, Abdulmotaleb El Saddik$^{4,6}$, Shijian Lu$^{1,\dagger}$, Eric Xing$^{4,5}$\\
$^{1}$Nanyang Technological University $^{2}$Waseda University $^{3}$Technical University of Denmark\\
$^{4}$Mohamed bin Zayed University of Artificial Intelligence\\ $^{5}$Carnegie Mellon University $^{6}$University of Ottawa \vspace{-15pt}
}

\twocolumn[{
\renewcommand\twocolumn[1][]{#1}
\maketitle
\begin{center}
    \centering
    \includegraphics[width=\linewidth]{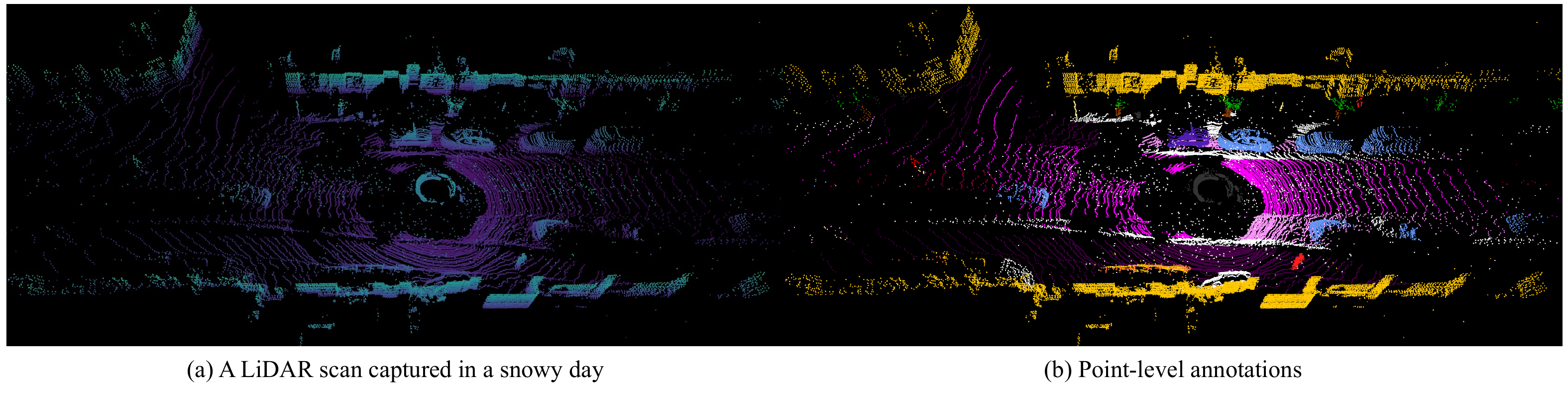}
    \vspace{-20pt}
    \captionof{figure}{We introduce SemanticSTF, an adverse-weather LiDAR point cloud dataset with dense point-level annotations that can be exploited for the study of point cloud semantic segmentation under all-weather conditions (including fog, snow, and rain). The graph on the left shows one scan sample captured on a snowy day, and the one on the right shows the corresponding point-level annotations.}
    \label{fig:motivation}
\end{center}
}]

\renewcommand{\thefootnote}{\fnsymbol{footnote}}
\begin{abstract}
\footnotetext{$^\dagger$ Corresponding author}
Robust point cloud parsing under all-weather conditions is crucial to level-5 autonomy in autonomous driving. However, how to learn a universal 3D semantic segmentation (3DSS) model is largely neglected as most existing benchmarks are dominated by point clouds captured under normal weather. We introduce SemanticSTF, an adverse-weather point cloud dataset that provides dense point-level annotations and allows to study 3DSS under various adverse weather conditions. We study all-weather 3DSS modeling under two setups:
1) domain adaptive 3DSS that adapts from normal-weather data to adverse-weather data; 
2) domain generalizable 3DSS that learns all-weather 3DSS models from normal-weather data.
Our studies reveal the challenge while existing 3DSS methods encounter adverse-weather data, showing the great value of SemanticSTF in steering the future endeavor along this very meaningful research direction. In addition, we design a domain randomization technique that alternatively randomizes the geometry styles of point clouds and aggregates their embeddings, ultimately leading to a generalizable model that can improve 3DSS under various adverse weather effectively. The SemanticSTF and related codes are available at \url{https://github.com/xiaoaoran/SemanticSTF}.
\end{abstract}

\section{Introduction}
3D LiDAR point clouds play an essential role in semantic scene understanding in various applications such as self-driving vehicles and autonomous drones. With the recent advance of LiDAR sensors, several LiDAR point cloud datasets~\cite{behley2019semantickitti,fong2022panoptic,xiao2022transfer} such as SemanticKITTI~\cite{behley2019semantickitti} have been proposed which greatly advanced the research in 3D semantic segmentation (3DSS)~\cite{hu2020randla, tang2020searching, zhu2021cylindrical} for the task of point cloud parsing. As of today, most existing point cloud datasets for outdoor scenes are dominated by point clouds captured under normal weather. However, 3D vision applications such as autonomous driving require reliable 3D perception under all-weather conditions including various adverse weather such as fog, snow, and rain. How to learn a weather-tolerant 3DSS model is largely neglected due to the absence of related benchmark datasets.

Although several studies~\cite{bijelic2020seeing,pitropov2021canadian} attempt to include adverse weather conditions in point cloud datasets, such as the STF dataset~\cite{bijelic2020seeing} that consists of LiDAR point clouds captured under various adverse weather, these efforts focus on object detection benchmarks and do not provide any point-wise annotations which are critical in various tasks such as 3D semantic and instance segmentation. To address this gap, we introduce \textit{SemanticSTF}, an adverse-weather point cloud dataset that extends the STF Detection Benchmark by providing point-wise annotations of 21 semantic categories, as illustrated in Fig.~\ref{fig:motivation}. Similar to STF, SemanticSTF captures four typical adverse weather conditions that are frequently encountered in autonomous driving including dense fog, light fog, snow, and rain.

SemanticSTF provides a great benchmark for the study of 3DSS and robust point cloud parsing under adverse weather conditions. Beyond serving as a well-suited test bed for examining existing fully-supervised 3DSS methods that handle adverse-weather point cloud data, SemanticSTF can be further  exploited to study two valuable weather-tolerant 3DSS scenarios: 1) domain adaptive 3DSS that adapts from normal-weather data to adverse-weather data, and 2) domain generalizable 3DSS that learns all-weather 3DSS models from normal-weather data. Our studies reveal the challenges faced by existing 3DSS methods while processing adverse-weather point cloud data, highlighting the significant value of SemanticSTF in guiding future research efforts along this meaningful research direction.

In addition, we design PointDR, a new baseline framework for the future study and benchmarking of all-weather 3DSS. Our objective is to learn robust 3D representations that can reliably represent points of the same category across different weather conditions while remaining discriminative across categories. However, robust all-weather 3DSS poses two major challenges: 
1) LiDAR point clouds are typically sparse, incomplete, and subject to substantial geometric variations and semantic ambiguity. These challenges are further exacerbated under adverse weather conditions, with many missing points and geometric distortions due to fog, snow cover, etc.
2) More noises are introduced under adverse weather due to snow flicks, rain droplets, etc. PointDR addresses the challenges with two iterative operations: 1) \textit{Geometry style randomization} that expands the geometry distribution of point clouds under various spatial augmentations; 2) \textit{Embedding aggregation} that introduces contrastive learning to aggregate the encoded embeddings of the randomly augmented point clouds. Despite its simplicity, extensive experiments over point clouds of different adverse weather conditions show that PointDR achieves superior 3DSS generalization performance.

The contribution of this work can be summarized in three major aspects. \textit{First}, we introduce SemanticSTF, a large-scale adverse-weather point cloud benchmark that provides high-quality point-wise annotations of 21 semantic categories. \textit{Second}, we design PointDR, a point cloud domain randomization baseline that can be exploited for future study and benchmarking of 3DSS under all-weather conditions. \textit{Third},  leveraging SemanticSTF, we benchmark existing 3DSS methods over two challenging tasks on domain adaptive 3DSS and domain generalized 3DSS. The benchmarking efforts lay a solid foundation for future research on this highly meaningful problem.

\section{Related Works}

\noindent\textbf{3D semantic segmentation} aims to assign point-wise semantic labels for point clouds. It has been developed rapidly over the past few years, largely through the development of various deep neural networks (DNNs) such as standard convolutional network for projection-based methods~\cite{wu2019squeezesegv2,milioto2019rangenet++,zhang2020polarnet,cortinhal2020salsanext,xiao2021fps}, multi-layer perceptron (MLP)-based networks~\cite{qi2017pointnet,qi2017pointnet,hu2020randla}, 3D voxel convolution-based networks~\cite{choy20194d,zhu2021cylindrical}, or hybrid networks~\cite{liu2019point,tang2020searching,cheng20212,zhang2020deep,xu2021rpvnet}. While existing 3DSS networks are mainly evaluated over normal weather point clouds, their performance for adverse weather point clouds is far under-investigated. The proposed SemanticSTF closes the gap and provides a solid ground for the study and evaluation of all-weather 3DSS. By enabling investigations into various new research directions, SemanticSTF represents a valuable tool for advancing the field.

\noindent\textbf{Vision recognition under adverse conditions.} 
Scene understanding under adverse conditions has recently attracted increasing attention due to the strict safety demand in various outdoor navigation and perception tasks. In 2D vision, several large-scale datasets have been proposed to investigate perceptions tasks in adverse visual conditions including localization~\cite{maddern20171}, detection~\cite{yu2020bdd100k}, and segmentation~\cite{sakaridis2021acdc}. On the other hand, learning 3D point clouds of adverse conditions is far under-explored due to the absence of comprehensive dataset benchmarks. The recently proposed datasets such as STF~\cite{bijelic2020seeing} and CADC~\cite{pitropov2021canadian} contain LiDAR point clouds captured under adverse weather conditions. However, these studies focus on the object detection task~\cite{hahner2021fog,hahner2022lidar} with bounding-box annotations, without providing any point-wise annotations. Our introduced SemanticSTF is the first large-scale dataset that consists of LiDAR point clouds in adverse weather conditions with high-quality dense annotations, to the best of our knowledge.

\noindent\textbf{Domain generalization}~\cite{blanchard2011generalizing,muandet2013domain} aims to learn a generalizable model from single or multiple related but distinct source domains where target data is inaccessible during model learning. It has been widely studied in 2D computer vision tasks~\cite{balaji2018metareg,li2018domain,huang2021fsdr,zhou2022domain} while few studies explore it in point cloud learning. Recently, \cite{lehner20223d} studies domain generalization for 3D object detection by deforming point clouds via vector fields. Differently, this work is the first attempt that explores domain generalization for 3DSS. 

\noindent\textbf{Unsupervised domain adaptation} is a method of transferring knowledge learned from a labeled source domain to a target domain by leveraging the unlabeled target data. It has been widely studied in 2D image learning~\cite{kang2019contrastive,ganin2015unsupervised,huang2021model,huang2021cross,guan2021domain,huang2022category} and 3D point clouds~\cite{su2020adapting,yang2021st3d,luo2021unsupervised,zhang2021srdan,xu2021spg,hahner2021fog,hahner2022lidar}.
Recently, domain adaptive 3D LiDAR segmentation has drawn increasing attention due to the challenge in point-wise annotation. Different UDA approaches have been designed to mitigate discrepancies across LiDAR point clouds of different domains. For example, \cite{wu2019squeezesegv2,zhao2021epointda} project point clouds into depth images and leverage 2D UDA techniques while~\cite{yi2021complete,xiao2022transfer,saltori2022cosmix,xiao2022polarmix} directly work in the 3D space. However, these methods either work for \textit{synthetic-to-real} UDA scenarios~\cite{wu2019squeezesegv2,xiao2022transfer} or \textit{normal-to-normal} point cloud adaptation~\cite{yi2021complete}, ignoring \textit{normal-to-adverse} adaptation which is highly practical in real applications. Our SemanticSTF dataset fills up this blank and will inspire more development of new algorithms for normal-to-adverse adaptation.

\section{The SemanticSTF Dataset}

\subsection{Background}
LiDAR sensors send out laser pulses and measure their flight time based on the echoes it receives from targets. The travel distance as derived from the time-of-flight and the registered angular information (between the LiDAR sensors and the targets) can be combined to compute the 3D coordinates of target surface which form point clouds that capture the 3D shape of the targets. However, the active LiDAR pulse system can be easily affected by the scattering media such as particles of rain droplets and snow~\cite{ryde2009performance, peynot2009towards,filgueira2017quantifying,heinzler2019weather}, leading to shifts of measured distances, variation of echo intensity, point missing, etc. Hence, point clouds captured under adverse weather usually have clear distribution discrepancy as compared with those collected under normal weather as illustrated in Fig.~\ref{fig:motivation}. However, existing 3DSS benchmarks are dominated by normal-weather point clouds which are insufficient for the study of universal 3DSS under all-weather conditions. To this end, we propose SemanticSTF, a point-wise annotated large-scale adverse-weather dataset that can be explored for the study of 3DSS and point cloud parsing under various adverse weather conditions.

\subsection{Data Selection and Split}
We collect SemanticSTF by leveraging the STF benchmark~\cite{bijelic2020seeing}, a multi-modal adverse-weather dataset that was jointly collected in Germany, Sweden, Denmark, and Finland. The data in STF have multiple modalities including LiDAR point clouds and they are collected under various adverse weather conditions such as snow and fog. However, STF provides bounding-box annotations only for the study of 3D detection tasks. In SemanticSTF, we manually selected 2,076 scans captured by a Velodyne HDL64 S3D LiDAR sensor from STF that cover various adverse weather conditions including 694 snowy, 637 dense-foggy, 631 light-foggy, and 114 rainy (all rainy LiDAR scans in STF).
During the selection, we pay special attention to the geographical diversity of the point clouds aiming for minimizing data redundancy.  We ignore the factor of daytime/nighttime since LiDAR sensors are robust to lighting conditions. We split SemanticSTF into three parts including 1,326 full 3D scans for training, 250 for validating, and 500 for testing. 
All three splits have approximately the same proportion of LiDAR scans of different adverse weathers.

\subsection{Data Annotation}
Point-wise annotation of LiDAR point clouds is an extremely laborious task due to several factors, such as 3D view changes, inconsistency between point cloud display and human visual perception, sweeping occlusion, point sparsity, etc. However, point-wise annotating of adverse-weather point clouds is even more challenging due to two new factors. \textit{First}, the perceived distance shifts under adverse weather often lead to various geometry distortions in the collected points which make them different from those collected under normal weather. This presents significant challenges for annotators who must recognize various objects and assign a semantic label to each point. \textit{Second}, LiDAR point clouds collected under adverse weather often contain a significant portion of invalid regions that consist of indiscernible semantic contents (e.g., thick snow cover) that make it difficult to identify the ground type. The existence of such invalid regions makes point-wise annotation even more challenging.

We designed a customized labeling pipeline to handle the annotation challenges while performing point-wise annotation of point clouds in SemanticSTF. Specifically, we first provide labeling instructions and demo annotations and train a team of professional annotators to provide point-wise annotations of a set of selected STF LiDAR scans. To achieve reliable high-quality annotations, the annotators leverage the corresponding 2D camera images and Google Street views as extra references while identifying the category of each point in this initial annotation process. After that, the annotators cross-check their initial annotations for identifying and correcting labeling errors. At the final stage, we engaged professional third parties who provide another round of annotation inspection and correction. 

Annotation of SemanticSTF is a highly laborious and time-consuming task. For instance, while labeling downtown areas with the most complex scenery, it took an annotator an average of 4.3 hours to label a single LiDAR scan. Labeling a scan captured in a relatively simpler scenery, such as a highway, also takes an average of 1.6 hours. In addition, an additional 30-60 minutes are required per scan for verification and correction by professional third parties. In total, annotating the entire SemanticSTF dataset takes over 6,600 man-hours.

While annotating SemanticSTF, we adopted the same set of semantic classes as in the widely-studied semantic segmentation benchmark, SemanticKITTI~\cite{behley2019semantickitti}. Specifically, we annotate the 19 evaluation classes of SemanticKITTI, which encompass most traffic-related objects in autonomous driving scenes. Additionally, following~\cite{sakaridis2021acdc}, we label points with indiscernible semantic contents caused by adverse weather (e.g. ground covered by snowdrifts) as \textit{invalid}. Furthermore, we label points that do not belong to the 20 categories or are indistinguishable as \textit{ignored}, which are not utilized in either training or evaluations. Detailed descriptions of each class can be found in the appendix.

\subsection{Data Statistics}

\begin{figure}[t]
    \centering
    \includegraphics[width=0.98\linewidth]{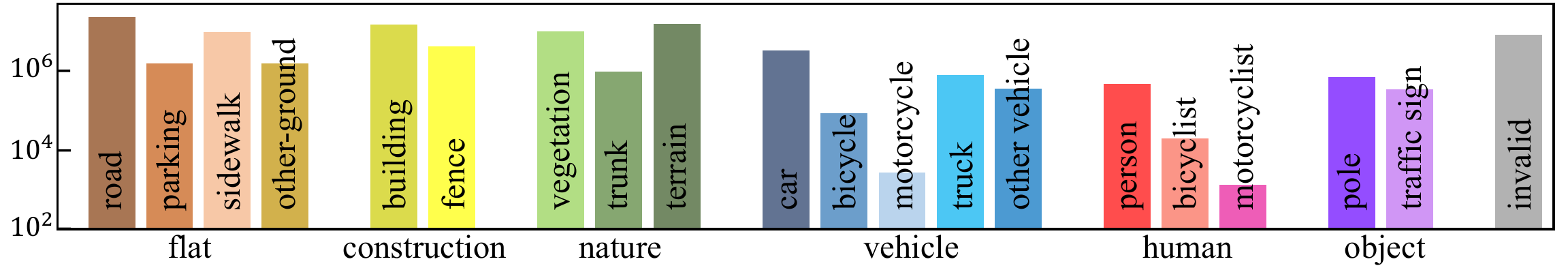}
    \caption{Number of annotated points per class in SemanticSTF.}
    \label{fig:class_dist}
\end{figure}

SemanticSTF consists of point-wise annotations of 21 semantic categories, and Fig.~\ref{fig:class_dist} shows the detailed statistics of the point-wise annotations. It can be seen that classes \textit{road}, \textit{sidewalk}, \textit{building}, \textit{vegetation}, and \textit{terrain} appear most frequently whereas classes \textit{motor}, \textit{motorcyclist}, and \textit{bicyclist} have clearly lower occurrence frequency. Such class imbalance is largely attributed to the various object sizes and unbalanced distribution of object categories in transportation scenes, and it is also very common in many existing benchmarks. Overall, the statistics and distribution of different object categories are similar to that of other 2D and 3D semantic segmentation benchmarks such as Cityscapes~\cite{cordts2016cityscapes}, ACDC~\cite{sakaridis2021acdc}, and SemanticKITTI~\cite{behley2019semantickitti}.

To the best of our knowledge, SemanticSTF is the first large-scale adverse-weather 3DSS benchmark that provides high-quality point-wise annotations. Table~\ref{tab:comparison} compares it with several existing point cloud datasets that have been widely adopted for the study of 3D detection and semantic segmentation. We can observe that existing datasets are either collected under normal weather conditions or collected for object detection studies with bounding-box annotations only. 3DSS benchmark under adverse weather is largely blank, mainly due to the great challenge in point-wise annotations of adverse-weather point clouds as described in previous subsections. From this sense, SemanticSTF fills up this blank by providing a large-scale benchmark and test bed which will be very useful to future research in universal 3DSS under all weather conditions.

\subsection{Data illustration}
\begin{figure*}[t]
    \centering
    \includegraphics[width=\linewidth]{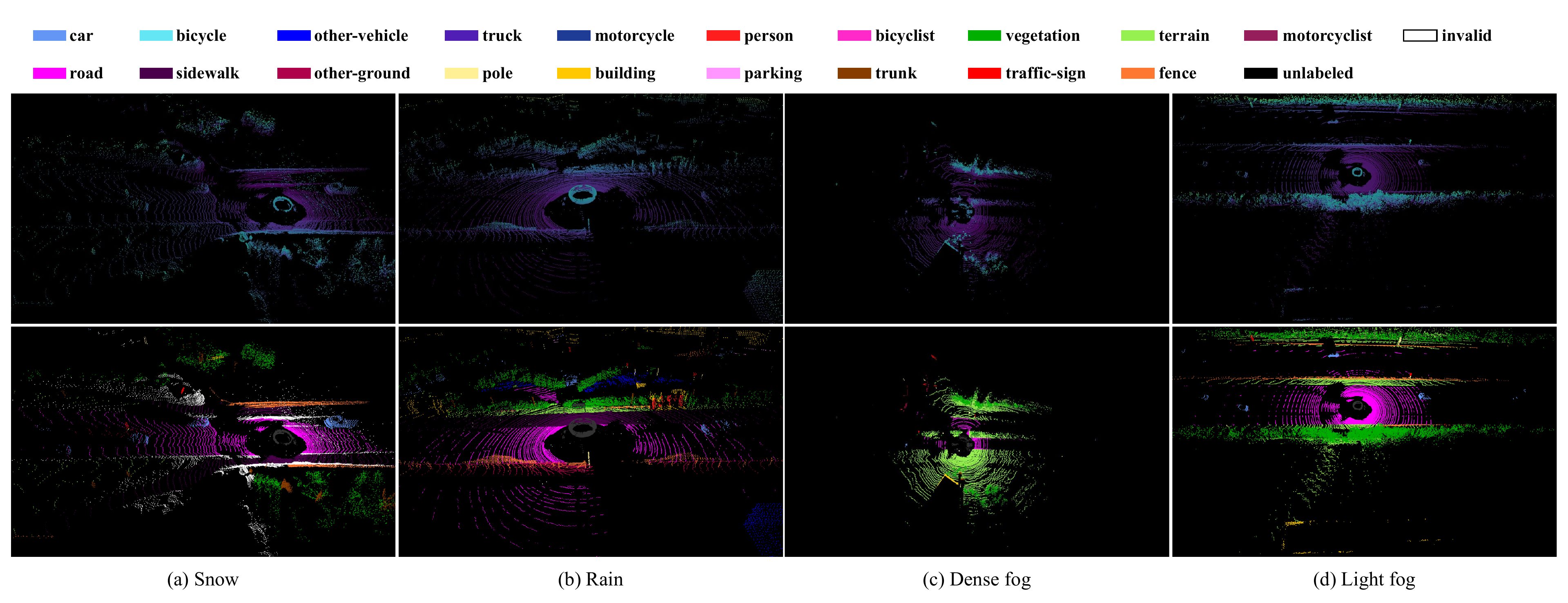}
    \vspace{-20pt}
    \caption{Examples of LiDAR point cloud scans captured under different adverse weather including snow, rain, dense fog, and light fog (the first row) and corresponding dense annotations in SemanticSTF (the second row).}
    \label{fig:semanticstf-examples}
    \vspace{-10pt}
\end{figure*}

Fig.~\ref{fig:semanticstf-examples} provides examples of point cloud scans captured under adverse weather conditions in SemanticSTF (in row 1) as well as the corresponding annotations (in row 2). Compared with normal-weather point clouds, point clouds captured under adverse weather exhibit four distinct properties: 1) Snow coverage and snowflakes under \textit{snowy} weather introduce many white points (labeled as ``invalid") as illustrated in Fig.~\ref{fig:semanticstf-examples}(a). The thick snow coverage may lead to object deformation as well; \textit{Rainy} conditions may cause specular reflection of laser signals from water on the ground and produce many noise points as shown in Fig.\ref{fig:semanticstf-examples}(b);
3) \textit{Dense fog} may greatly reduce the working range of LiDAR sensors, leading to small spatial distribution of the collected LiDAR points as illustrated in Fig.~\ref{fig:semanticstf-examples}(c); 4) Point clouds under \textit{light fog} have similar characteristics as normal-weather point clouds as illustrated in Fig.~\ref{fig:semanticstf-examples}(d). The distinct properties of point clouds under different adverse weather introduce different types of domain shift from normal-weather point clouds which complicate 3DSS greatly as discussed in Section~\ref{sec.exp}. They also verify the importance of developing universal 3DSS models that can perform well under all weather conditions.

\begin{table}[t]
    \begin{footnotesize}
    \setlength{\tabcolsep}{2pt}
    \centering
    \begin{tabular}{rcccccc}
    \hline
        Dataset &  \#Cls & Type & Annotation  & Fog & Rain & Snow\\
    \hline
        KITTI~\cite{geiger2013vision} & 8 & real & bounding box & \xmark & \xmark & \xmark\\
        nuScenes~\cite{caesar2020nuscenes} & 23 & real & bounding box & \xmark & \xmark & \xmark\\
        Waymo~\cite{sun2020waymo} &  4 & real & bounding box & \xmark & \xmark & \xmark\\
        STF~\cite{bijelic2020seeing} & 5 & real & bounding box & \cmark & \cmark & \cmark\\
    \hline
        SemanticKITTI~\cite{behley2019semantickitti} & 25 & real & point-wise & \xmark & \xmark & \xmark\\
        nuScenes-LiDARSeg~\cite{fong2022panoptic} & 32 & real & point-wise & \xmark & \xmark & \xmark\\
        Waymo-LiDARSeg~\cite{sun2020waymo} & 21 & real & point-wise & \xmark & \xmark & \xmark\\
        SynLiDAR~\cite{xiao2022transfer} & 32 & synth. & point-wise & \xmark & \xmark & \xmark\\
        SemanticSTF (ours) & 21 & real & point-wise & \cmark & \cmark & \cmark\\
    \hline
    \end{tabular}
    \caption{Comparison of SemanticSTF against existing outdoor LiDAR benchmarks. \#Cls means the class number.}
    \label{tab:comparison}
    \end{footnotesize}
\end{table}

\section{Point Cloud Domain Randomization}

Leveraging SemanticSTF, we explore domain generalization (DG) for semantic segmentation of LiDAR point clouds under all weather conditions. Specifically, we design PointDR, a domain randomization technique that helps to train a generalizable segmentation model from normal-weather point clouds that can work well for adverse-weather point clouds in SemanticSTF.

\subsection{Problem Definition}\label{subsec.preliminaries}
Given labeled point clouds of a source domain $\mathcal{S}={\{S_k=\{x_k,y_k\}\}}^K_{k=1}$ where $x$ represents a LiDAR point cloud scan and $y$ denotes its point-wise semantic annotations, the goal of domain generalization is to learn a segmentation model $F$ by using the source-domain data only that can perform well on
point clouds from an unseen target domain $\mathcal{T}$.
We consider a 3D point cloud segmentation model $F$ that consists of a feature extractor $E$ and a classifier $G$.
Note under the setup of domain generalization, target data will not be accessed in training as they could be hard and even impossible to acquire at the training stage.

\subsection{Point Cloud Domain Randomization}\label{subsec.pointdr}

\begin{figure}[t]
    \centering
    \includegraphics[width=\linewidth]{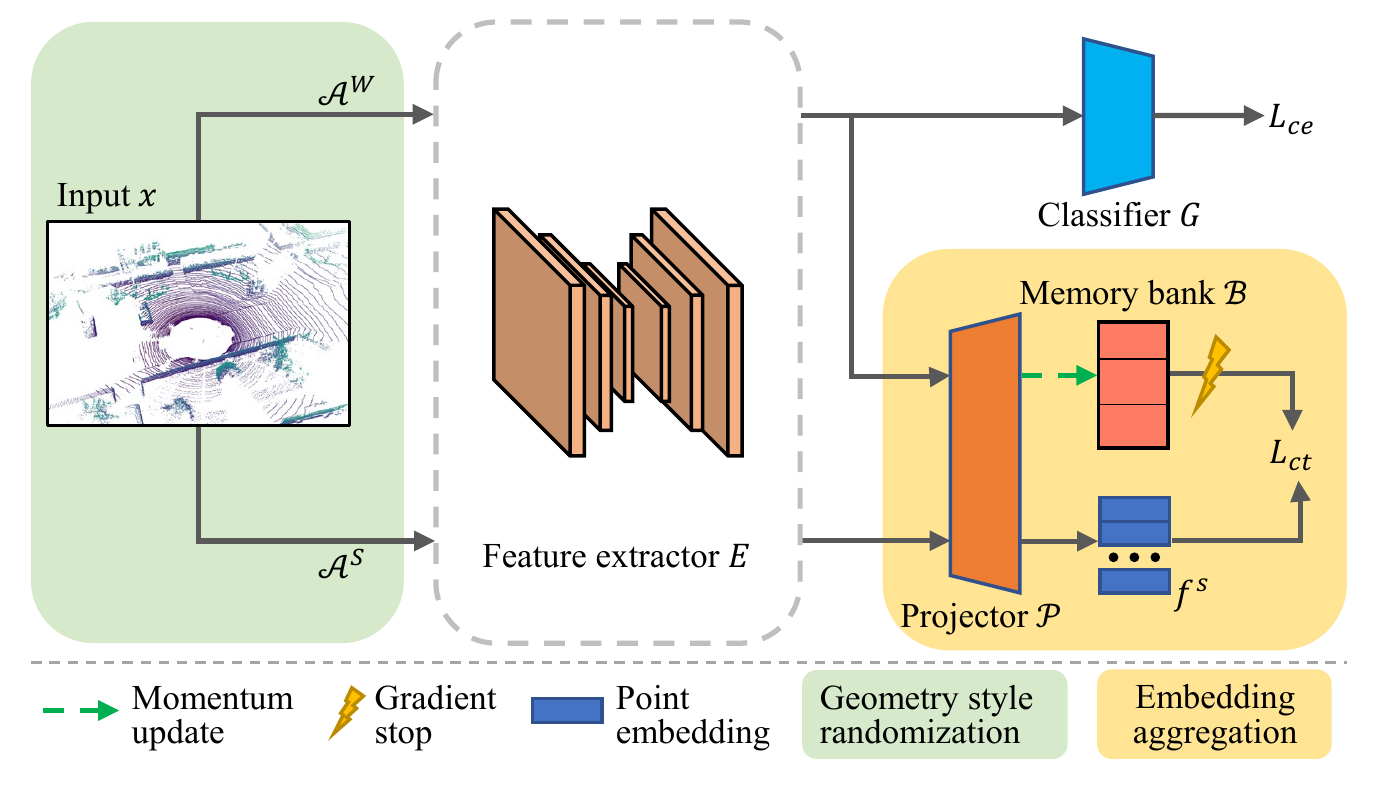}
    \vspace{-20pt}
    \caption{The framework of our point cloud randomization method (PointDR): \textit{Geometry style randomization} creates different point cloud views with various spatial perturbations while \textit{embedding aggregation} encourages the feature extractor to aggregate randomized point embeddings to learn perturbation-invariant representations, ultimately leading to a generalizable segmentation model.}
    \vspace{-10pt}
    \label{fig:pipeline}
\end{figure}

Inspired by domain randomization studies in 2D computer vision research~\cite{tobin2017domain,tremblay2018training}, we explore how to employ domain randomization for learning domain generalizable models for point Specifically, we design PointDR, a point cloud randomization technique that consists of two complementary designs including \textit{geometry style randomization} and \textit{embedding aggregation} as illustrated in Fig.~\ref{fig:pipeline}.

\noindent\textbf{Geometry style randomization} aims to enrich the geometry styles and expand the distribution of training point cloud data. Given a point-cloud scan $x$ as input, we apply weak and strong spatial augmentation to obtain two copies of $x$ including a weak-view $x^w=\mathcal{A}^W(x)$ and a strong-view $x^s=\mathcal{A}^S(x)$. For the augmentation schemes of $\mathcal{A}^W$, we follow existing supervised learning methods~\cite{tang2020searching} and adopt the simple random rotation and random scaling. While for the augmentation schemes of $\mathcal{A}^S$, we further adopt random dropout, random flipping, random noise perturbation, and random jittering on top of $\mathcal{A}^W$ to obtain a more diverse and complex copy of the input point cloud scan $x$.

\noindent\textbf{Embedding aggregation} aims to aggregate encoded embeddings of randomized point clouds for learning domain-invariant representations. We adopt contrastive learning~\cite{he2020momentum} as illustrated in Fig.~\ref{fig:pipeline}. 
Given the randomized point clouds $x^w$ and $x^s$, we first feed them into the \textit{feature extractor} $E$ and a \textit{projector} $\mathcal{P}$ (a two-layer MLP) which outputs normalized point feature embeddings $f^w$ and $f^s$, respectively ($f=\mathcal{P}(E(x))$). 
$\overline{f}^w_C\in\mathbb{R}^{D\times C}$ ($D$: feature dimension; $C$: number of semantic classes) is then derived by class-wise averaging the feature embeddings $f^w$ in a batch, which is stored in a memory bank $\mathcal{B}\in\mathbb{R}^{D\times C}$ that has no backpropagation and is momentum updated by iterations (i.e., $\mathcal{B}\leftarrow m\times \mathcal{B} + (1-m)\times \overline{f}^w_C$ with a momentum coefficient $m$).
Finally, we employ each point feature embedding $f^s_i$ of the strong-view $f^s$ as query and feature embeddings in $\mathcal{B}$ as keys for contrastive learning, where the key sharing the same semantic class as the query is positive key $\mathcal{B}_+$ and the rest are negative keys. The contrastive loss is defined as

\renewcommand\arraystretch{1.0}
\setlength{\tabcolsep}{0.7mm}{
\begin{table*}[t]
\centering
\begin{footnotesize}
\begin{tabular}{l|ccccccccccccccccccc|cccc|c}
 \toprule
  Methods & \rotatebox{90}{car} & \rotatebox{90}{bi.cle} & \rotatebox{90}{mt.cle} & \rotatebox{90}{truck} & \rotatebox{90}{oth-v.} & \rotatebox{90}{pers.} & \rotatebox{90}{bi.clst} & \rotatebox{90}{mt.clst} & \rotatebox{90}{road} & \rotatebox{90}{parki.} & \rotatebox{90}{sidew.} & \rotatebox{90}{oth-g.} & \rotatebox{90}{build.} & \rotatebox{90}{fence} & \rotatebox{90}{veget.} & \rotatebox{90}{trunk} & \rotatebox{90}{terra.} & \rotatebox{90}{pole} & \rotatebox{90}{traf.} & \rotatebox{90}{D-fog} & \rotatebox{90}{L-fog} &  \rotatebox{90}{Rain} & \rotatebox{90}{Snow} & mIoU \\
  \midrule
   Oracle & 89.4 & 42.1 & 0.0 & 59.9 & 61.2 & 69.6 & 39.0 & 0.0 & 82.2 & 21.5 & 58.2 & 45.6 & 86.1 & 63.6 & 80.2 & 52.0 & 77.6 & 50.1 & 61.7 & 51.9 & 54.6 & 57.9 & 53.7 & 54.7\\ 
  \midrule
  \multicolumn{25}{c}{SemanticKITTI$\rightarrow$SemanticSTF}\\
  \midrule
  Baseline & 55.9 & 0.0 & 0.2 & 1.9 & 10.9 & 10.3 & \textbf{6.0} & 0.0 & 61.2 & 10.9 & 32.0 & 0.0 & 67.9 & 41.6 & 49.8 & 27.9 & 40.8 & \textbf{29.6} & 17.5 & 29.5 & 26.0 & 28.4 & 21.4 & 24.4 \\
  Dropout~\cite{srivastava2014dropout} &  62.1 & 0.0 & \textbf{15.5} & 3.0 & \textbf{11.5} & 5.4 & 2.0 & 0.0 & 58.4 & 12.8 & 26.7 & 1.1 & 72.1 & 43.6 & 52.9 & \textbf{34.2} & 43.5 & 28.4 & 15.5 & 29.3 & 25.6 & 29.4 & 24.8 & 25.7 \\
  Perturbation & \textbf{74.4} & 0.0 & 0.0 & \textbf{23.3} & 0.6 & 19.7 & 0.0 & 0.0 & 60.3 & 10.8 & 33.9 & 0.7 & 72.0 & 45.2 & 58.7 & 17.5 & 42.4 & 22.1 & 9.7 & 26.3 & 27.8 & 30.0 & 24.5 & 25.9 \\
  PolarMix~\cite{xiao2022polarmix} & 57.8 & \textbf{1.8} & 3.8 & 16.7 & 3.7 & 26.5 & 0.0 & \textbf{2.0} & 65.7 & 2.9 & 32.5 & 0.3 & 71.0 & \textbf{48.7} & 53.8 & 20.5 & 45.4 & 25.9 & 15.8 & 29.7 & 25.0 & 28.6 & 25.6 & 26.0 \\
  MMD~\cite{li2018domain} & 63.6 & 0.0 & 2.6 & 0.1 & 11.4 & \textbf{28.1} & 0.0 & 0.0 & \textbf{67.0} & \textbf{14.1} & 37.9 & 0.3 & 67.3 & 41.2 & 57.1 & 27.4 & 47.9 & 28.2 & 16.2 & 30.4 & 28.1 & \textbf{32.8} & 25.2 & 26.9 \\
  PCL~\cite{yao2022pcl} & 65.9 & 0.0 & 0.0 & 17.7 & 0.4 & 8.4 & 0.0 & 0.0 & 59.6 & 12.0 & 35.0 & \textbf{1.6} & \textbf{74.0} & 47.5 & \textbf{60.7} & 15.8 & \textbf{48.9} & 26.1 & \textbf{27.5} & 28.9 & 27.6 & 30.1 & 24.6 & 26.4 \\
  PointDR (Ours) &  67.3 & 0.0 & 4.5 & 19.6 & 9.0 & 18.8 & 2.7 & 0.0 & 62.6 & 12.9 & \textbf{38.1} & 0.6 & 73.3 & 43.8 & 56.4 & 32.2 & 45.7  & 28.7 & 27.4 & \textbf{31.3} & \textbf{29.7} & 31.9 & \textbf{26.2} & \textbf{28.6} \\
  \midrule
  \multicolumn{25}{c}{SynLiDAR$\rightarrow$SemanticSTF}\\
  \midrule
  Baseline & 27.1& \textbf{3.0}& 0.6& 15.8& 0.1& 25.2& 1.8& 5.6& 23.9& 0.3& 14.6& 0.6& 36.3& 19.9 & 37.9& 17.9& 41.8& 9.5& 2.3 & 16.9 & 17.2 & 17.2 & 11.9 & 15.0 \\
  Dropout~\cite{srivastava2014dropout} & 28.0 & \textbf{3.0} & 1.4 & 9.6 & 0.0 & 17.1 & 0.8 & 0.7 & \textbf{34.2} & 6.8 & 19.1 & 0.1 & 35.5 & 19.1 & 42.3 & 17.6 & 36.0 & 14.0 & 2.8 & 15.3 & 16.6 & 20.4 & 14.0 & 15.2 \\
  Perturbation & 27.1 & 2.3 & 2.3 & 16.0 & 0.1 & 23.7 & 1.2 & 4.0 & 27.0 & 3.6 & 16.2 & 0.8 & 29.2 & 16.7 & 35.3 & 22.7 & 38.3 & \textbf{17.9} & 5.1 & 16.3 & 16.7 & 19.3 & 13.4 & 15.2 \\
  PolarMix~\cite{xiao2022polarmix} & \textbf{39.2} & 1.1 & 1.2 & 8.3 & \textbf{1.5} & 17.8 & 0.8 & 0.7 & 23.3 & 1.3 & 17.5 & 0.4 & 45.2 & 24.8 & \textbf{46.2} & 20.1 & 38.7 & 7.6 & 1.9 & 16.1 & 15.5 & 19.2 & 15.6 & 15.7 \\
  MMD~\cite{li2018domain} &  25.5 & 2.3 & 2.1 & 13.2 & 0.7 & 22.1 & 1.4 & 7.5 & 30.8 & 0.4 & 17.6 & 0.2 & 30.9 & 19.7 & 37.6 & 19.3 & \textbf{43.5} & 9.9 & 2.6 & 17.3 & 16.3 & 20.0 & 12.7  & 15.1 \\
  PCL~\cite{yao2022pcl} &  30.9 &  0.8 &  1.4 &  10.0 &  0.4 &  23.3 &  \textbf{4.0} &  \textbf{7.9} &  28.5 &  1.3 &  \textbf{17.7} &  \textbf{1.2} &  39.4 &  18.5 &  40.0 &  16.0 &  38.6 &  12.1 &  2.3 & 17.8 & 16.7 & 19.3 & 14.1 & 15.5 \\
   PointDR (Ours) & 37.8 & 2.5 & \textbf{2.4} & \textbf{23.6} & 0.1 & \textbf{26.3} & 2.2 & 3.3 & 27.9 & \textbf{7.7} & 17.5 & 0.5 & \textbf{47.6} & \textbf{25.3} & 45.7 & \textbf{21.0} & 37.5 & \textbf{17.9} & \textbf{5.5} & \textbf{19.5} & \textbf{19.9} & \textbf{21.1} & \textbf{16.9} & \textbf{18.5} \\
    \bottomrule
    \end{tabular}
    \caption{Experiments on domain generalization with SemanticKITTI~\cite{behley2019semantickitti} or SynLiDAR~\cite{xiao2022transfer} as source and SemanticSTF as target.}
    \label{tab:dg_stf}
  \end{footnotesize}
\end{table*}
}

\begin{equation}
    \mathcal{L}_{ct}=\frac{1}{N}\sum_{i=1}^N-\log{\frac{\exp{(f^s_i\mathcal{B}_+}/\tau)}{\sum_{j=1}^C\exp{(f^s_i\mathcal{B}_j/\tau)}}}
    \label{eq.contrast}
\end{equation}
where $\tau$ is a temperature hyper-parameter~\cite{wu2018unsupervised}. Note there is no back-propagation for the ``ignore" class in optimizing the contrastive loss.

Contrastive learning pulls point feature embeddings of the same classes closer while pushing away point feature embeddings of different classes. Therefore, optimizing the proposed contrastive loss will aggregate randomized point cloud features and learn perturbation-invariant representations, ultimately leading to a robust and generalizable segmentation model. The momentum-updated memory bank provides feature prototypes of each semantic class for more robust and stable contrastive learning.

Combining the supervised cross-entropy loss $\mathcal{L}_{ce}$ for weakly-augmented point clouds in Eq.~\ref{eq.contrast}, the overall training objective of  PointDR can be formulated by:
\begin{equation}
    \mathcal{L}_{\mathrm{PointDR}}=\mathcal{L}_{ce}+\lambda_{ct}\mathcal{L}_{ct}
\end{equation}

\section{Evaluation of Semantic Segmentation}\label{sec.exp}

SemanticSTF can be adopted for benchmarking different learning setups and network architectures on point cloud segmentation. We perform experiments over two typical learning setups including domain generalization and unsupervised domain adaptation. In addition, we evaluate several state-of-the-art point-cloud segmentation networks to examine their generalization capabilities.

\subsection{Domain Generalization}
We first study domain generalizable point cloud segmentation. For DG, we can only access an annotated source domain during training and the trained model is expected to generalize well to \textit{unseen} target domains. Leveraging SemanticSTF, we build two DG benchmarks and examine how PointDR helps learn a universal 3DSS model that can work under different weather conditions.

The first benchmark is \textit{SemanticKITTI~\cite{behley2019semantickitti} $\rightarrow$ SemanticSTF} where SemanticKITTI is a large-scale real-world 3DSS dataset collected under normal weather conditions. This benchmark serves as a solid testing ground for evaluating domain generalization performance from normal to adverse weather conditions.
The second benchmark is \textit{SynLiDAR~\cite{xiao2022transfer} $\rightarrow$ SemanticSTF} where SynLiDAR is a large-scale synthetic 3DSS dataset. The motivation of this benchmark is that learning a universal 3DSS model from synthetic point clouds that can work well across adverse weather is of high research and application value considering the challenges in point cloud collection and annotation. Note this benchmark is more challenging as the domain discrepancy comes from both normal-to-adverse weather distribution shift and synthetic-to-real distribution shift.

\noindent\textbf{Setup.}
We use all 19 evaluating classes of SemanticKITTI in both domain generalization benchmarks. The category of \textit{invalid} in SemanticSTF is mapped to the \textit{ignored} since SemanticKITTI and SynLiDAR do not cover this category. 
We adopt MinkowskiNet~\cite{choy20194d} (with TorchSparse library~\cite{tang2020searching}) as the backbone model, which is a sparse convolutional network that provides state-of-the-art performance with decent efficiency. We adopt the evaluation metrics of Intersection over the Union (IoU) for each segmentation class and the mean IoU (mIoU) over all classes. All experiments are run over a single NVIDIA 2080Ti (11GB). More implementation details are provided in the appendix.

\noindent\textbf{Baseline Methods.} 
Since domain generalizable 3DSS is far under-explored, there is little existing baseline that can be directly adopted for benchmarking. We thus select two closely related approaches as baseline to evaluate the proposed PointDR. The first approach is data augmentation and we select three related augmentation methods including \textit{Dropout}~\cite{srivastava2014dropout} that randomly drops out points to simulate LiDAR points missing in adverse weather, \textit{Noise perturbation} that adds random points in the 3D space to simulate noise points as introduced by particles like falling snow, and \textit{PolarMix}~\cite{xiao2022polarmix} that mixes point clouds of different sources for augmentation. The second approach is to adapt 2D domain generalization methods for 3DSS. We select two 2D domain generalization methods including the widely studied \textit{MMD}~\cite{li2018domain} and the recently proposed \textit{PCL}~\cite{yao2022pcl}.

\setlength{\tabcolsep}{3.0mm}{
\begin{table}[t]
    \centering
    \begin{footnotesize}
    \begin{tabular}{l|ccc|c}
    \toprule
        Method & $\mathcal{L}_{ce}$ & $\mathcal{L}_{ct}$ & $\mathcal{B}$ & mIoU\\
    \midrule
        Baseline & \checkmark & & & 24.4 \\
        PointDR-CT & \checkmark & \checkmark & & 27.4\\
        PointDR & \checkmark & \checkmark & \checkmark & 28.6\\
    \bottomrule
    \end{tabular}
    \caption{Ablation study of PointDR over domain generalized segmentation task SemanticKITTI$\rightarrow$SemanticSTF.}
    \label{tab:ablation}
    \end{footnotesize}
\end{table}
}

\setlength{\tabcolsep}{1.2mm}{
\begin{table*}[t]
\centering
\begin{footnotesize}
\begin{tabular}{l|ccccccccccccccccccc|c}
    \toprule
    Methods & \rotatebox{90}{car} & \rotatebox{90}{bi.cle} & \rotatebox{90}{mt.cle} & \rotatebox{90}{truck} & \rotatebox{90}{oth-v.} & \rotatebox{90}{pers.} & \rotatebox{90}{bi.clst} & \rotatebox{90}{mt.clst} & \rotatebox{90}{road} & \rotatebox{90}{parki.} & \rotatebox{90}{sidew.} & \rotatebox{90}{oth-g.} & \rotatebox{90}{build.} & \rotatebox{90}{fence} & \rotatebox{90}{veget.} & \rotatebox{90}{trunk} & \rotatebox{90}{terra.} & \rotatebox{90}{pole} & \rotatebox{90}{traf.} & mIoU\\
    \midrule
    Oracle & 89.4 & 42.1 & 0.0 & 59.9 & 61.2 & 69.6 & 39.0 & 0.0 & 82.2 & 21.5 & 58.2 & 45.6 & 86.1 & 63.6 & 80.2 & 52.0 & 77.6 & 50.1 & 61.7 & 54.7\\ 
    \midrule
    Source-only & 64.8 & 0.0 & 0.0 & 13.8 & 1.8 & 5.0 & 2.1 & 0.0 & 62.7 & 7.5 & 34.0 & 0.0 & 66.7 & 36.2 & 53.9 & 31.3 & 44.3 & 24.0 & 14.2 & 24.3 \\
    ADDA~\cite{tzeng2017adda} & 65.6 & 0.0 & 0.0 & 21.0 & 1.3 & 2.8 & 1.3 & \textbf{16.7} & 64.7 & 1.2 & 35.4 & 0.0 & 66.5 & 41.8 & \textbf{57.2} & 32.6 & 42.2 & 23.3 & 26.4 & 26.3 \\
    Ent-Min~\cite{vu2019advent} & 69.2 & 0.0 & 10.1 & 31.0 & 5.3 & 2.8 & \textbf{2.6} & 0.0 & \textbf{65.9} & 2.6 & 35.7 & 0.0 & \textbf{72.5} & \textbf{42.8} & 52.4 & 32.5 & \textbf{44.7} & 24.7 & 21.1 & 27.2 \\
    Self-training~\cite{zou2019confidence} & \textbf{71.5} & 0.0 & 10.3 & \textbf{33.1} & 7.4 & 5.9 & 1.3 & 0.0 & 65.1 & 6.5 & \textbf{36.6} & 0.0 & 67.8 & 41.3 & 51.7 & \textbf{32.9} & 42.9 & 25.1 & 25.0 & 27.6  \\
    CoSMix~\cite{saltori2022cosmix} & 65.0 & \textbf{1.7} & \textbf{22.1} & 25.2 & \textbf{7.7} & \textbf{33.2} & 0.0 & 0.0 & 64.7 & \textbf{11.5} & 31.1 & \textbf{0.9} & 62.5 & 37.8 & 44.6 & 30.5 & 41.1 & \textbf{30.9} & \textbf{28.6} & \textbf{28.4}  \\
    \bottomrule
    \end{tabular}
    \vspace{-5pt}
    \caption{Comparison of state-of-the-art domain adaptation methods on SemanticKITTI$\rightarrow$SemanticSTF adaptation. SemanticKITTI serves as the source domain and the entire SemanticSTF including all four weather conditions serves as the target domain. 
    }
    \label{tab:uda_kitti2stf}
    \end{footnotesize}
\end{table*}
}

\noindent\textbf{Results.} 
Table~\ref{tab:dg_stf} shows experimental results over the validation set of SemanticSTF.
For both benchmarks, the \textit{Baseline} is a source-only model that is trained by using the training data of SemanticKITTI or SynLiDAR. We can see that the \textit{Baseline} achieves very low mIoU while evaluated over the validation set of SemanticSTF, indicating the large domain discrepancy between point clouds of normal and adverse weather conditions. In addition, all three data augmentation methods improve the model generalization consistently but the performance gains are limited especially for the challenging benchmark SynLiDAR$\rightarrow$ SemanticSTF. The two 2D generalization methods both help SemanticKITTI $\rightarrow$ SemanticSTF clearly but show very limited improvement over SynLiDAR $\rightarrow$ SemanticSTF. The proposed PointDR achieves the best generalization consistently across both benchmarks, demonstrating its superior capability to learn perturbation-invariant point cloud representations and effectiveness while handling all-weather 3DSS tasks.

We also evaluate the compared domain generalization methods over each individual adverse weather condition as shown in Table~\ref{tab:dg_stf}. It can be observed that the three data augmentation methods work for data captured in rainy and snowy weather only. The 2D generalization method MMD shows clear effectiveness for point clouds under dense fog and rain while PCL works for point clouds under rainy and snowy weather instead. 
We conjecture that the performance variations are largely attributed to the different properties of point clouds captured under different weather conditions. For example, more points are missing in rain while object points often deform due to the covered snow (more illustrations are provided in the appendix). Such data variations lead to different domain discrepancies across weather which further leads to different performances of the compared methods. As PointDR learns perturbation-tolerant representations, it works effectively across different adverse weather conditions. We also provide qualitative results, please refer to the appendix for details.

\noindent\textbf{Ablation study}.
We study different PointDR designs to examine how they contribute to the overall generalization performance. As Table~\ref{tab:ablation} shows, we report three models over the benchmark “SemanticKITTI $\rightarrow$ SemanticSTF”: 1) \textit{Baseline} that is trained with $\mathcal{L}_{ce}$. 2) \textit{PointDR-CT} that is jointly trained with $\mathcal{L}_{ce}$ and $\mathcal{L}_{ct}$ without using the memory bank $\mathcal{B}$. 3) The complete \textit{PointDR} that is trained with $\mathcal{L}_{ce}$, $\mathcal{L}_{ct}$ and the memory bank $\mathcal{B}$. 
We evaluate the three models over the validation set of SemanticSTF and Table~\ref{tab:ablation} shows experimental results. We can see that the \textit{Baseline} performs poorly at 24.4\% due to clear domain discrepancy between point clouds of normal weather and adverse weather. Leveraging the proposed contrastive loss, $\mathcal{L}_{ct}$ achieves clearly better performance at 27.4\%, indicating that learning perturbation-invariance is helpful for universal LiDAR segmentation of all-weather conditions. On top of that, introducing the momentum-updated memory bank $\mathcal{B}$ further improves the segmentation performance at 28.6\%.  This is because the feature embeddings in $\mathcal{B}$ serve as the class prototypes which help the optimization of the segmentation network, finally leading to more robust representations of 3DSS that perform better over adverse weather point clouds.

\subsection{Domain Adaptation}
 We also study SemanticSTF over a domain adaptive point cloud segmentation benchmark SemanticKITTI $\rightarrow$ SemanticSTF. Specifically, we select four representative UDA methods including ADDA~\cite{tzeng2017adda}, entropy minimization (Ent-Min)~\cite{vu2019advent}, self-training~\cite{zou2019confidence}, and CoSMix~\cite{saltori2022cosmix} for adaptation from the source SemanticKITTI~\cite{behley2019semantickitti} toward the target SemanticSTF. Following the state-of-the-art~\cite{xiao2022transfer,saltori2022cosmix,xiao2022polarmix} on synthetic-to-real adaptation, we adopt MinkowskiNet~\cite{choy20194d} as the segmentation backbone for all compared methods. Table~\ref{tab:uda_kitti2stf} shows experimental results over the validation set of SemanticSTF. We can see that all UDA methods outperform the \textit{Source-only} consistently under the normal-to-adverse adaptation setup. At the other end, the performance gains are still quite limited, showing the great improvement space along domain adaptive 3DSS from normal to adverse weather conditions.

In addition, we examined the adaptability of the four UDA methods in relation to each individual adverse weather condition. Specifically, we trained each of the four methods for adaptation from SemanticKITTI to SemanticSTF data for each adverse weather condition. Table~\ref{tab:da_by_weather} shows the experimental results over the validation set of SemanticSTF. We can see all four methods outperform the \textit{Source-only} method under \textit{Dense-fog} and \textit{Light-fog}, demonstrating their effectiveness in mitigating domain discrepancies. However, for \textit{rain} and \textit{Snow}, only CoSMix achieved marginal performance gains while the other three UDA methods achieved limited performance improvements. We conjecture that snow and rain introduce large deformations on object surfaces or much noise, making adaptation from normal to adverse weather more challenging. CoSMix works in the input space by directly mixing source and target points, allowing it to perform better under heavy snow and rain which have larger domain gaps. However, all methods achieved relatively low segmentation performance, indicating the significance of our research and the large room for improvement in our constructed benchmarks.

\setlength{\tabcolsep}{2.0mm}{
\begin{table}[t]
    \begin{footnotesize}
    \centering
    \begin{tabular}{l|cccc}
      \toprule
      Method & Dense-fog & Light-fog & Rain & Snow \\
      \midrule
      Source-Only & 26.9 & 25.2 & 27.7 & 23.5 \\
      ADDA~\cite{tzeng2017adda} & 31.5 & 27.9 & 27.4 & 23.4\\
      Ent-Min~\cite{vu2019advent} & 31.4 & 28.6 & 30.3 & 24.9\\
      Self-training~\cite{zou2019confidence} & \textbf{31.8} & 29.3 & 27.9 & 25.1\\
      CoSMix~\cite{saltori2022cosmix} & 31.6 & \textbf{30.3} &\textbf{ 33.1} & \textbf{32.9}\\
      \bottomrule
    \end{tabular}
    \vspace{-5pt}
    \caption{Comparison of state-of-the-art domain adaptation methods on SemanticKITTI$\rightarrow$SemanticSTF adaptation for individual adverse weather conditions. We train a separate model for each weather-specific subset of SemanticSTF and evaluate the trained model on the weather condition it has been trained for.}
    \label{tab:da_by_weather}
    \end{footnotesize}
\end{table}
}

\subsection{Network Models vs All-Weather 3DSS}

We also study how different 3DSS network architectures generalize when they are trained with normal-weather point clouds and evaluated over SemanticSTF. Specifically, we select five representative 3DSS networks~\cite{hu2020randla,cortinhal2020salsanext, tang2020searching,zhu2021cylindrical} that have been widely adopted in 3D LiDAR segmentation studies. In the experiments, each selected network is first pre-trained with SemanticKITTI~\cite{behley2019semantickitti} and then evaluated over the validation set of SemanticSTF. We directly use the officially released code and the pre-trained weights for evaluation.
Table~\ref{tab:pretrained} shows experimental results. We can observe that the five pre-trained models perform very differently though they all achieve superior segmentation over SemanticKITTI. Specifically, RandLA-Net~\cite{hu2020randla}, SPVCNN~\cite{tang2020searching}, and SPVNAS~\cite{tang2020searching} perform clearly better than SalsaNext~\cite{cortinhal2020salsanext} and Cylinder3D~\cite{zhu2021cylindrical}. In addition, none of the five pre-trained models perform well, verifying the clear domain discrepancy between point clouds of normal and adverse weather conditions. The experiments further indicate the great value of SemanticSTF in the future exploration of robust point cloud parsing under all weather conditions.
In addition, the supervised performance of these 3DSS networks over SemanticSTF is provided in the appendix.

\setlength{\tabcolsep}{3.0mm}{
\begin{table}[t]
    \begin{footnotesize}
    \centering
    \begin{tabular}{l|ccccc}
        \toprule
        3DSS Model &  D-fog & L-fog &  Rain & Snow & All \\
        \midrule
        RandLA-Net~\cite{hu2020randla} & 26.5& 26.0 & 25.1 & 22.7 & 25.3 \\
        SalsaNext~\cite{cortinhal2020salsanext} & 16.0 & 9.6 & 7.8 & 3.5 & 9.1\\
        SPVCNN~\cite{tang2020searching}  & 30.4 & 22.8 & 21.7 & 18.3 & 22.4\\
        SPVNAS~\cite{tang2020searching} & 25.5 & 18.3 & 17.0 & 13.0 & 18.0 \\
        Cylinder3D~\cite{zhu2021cylindrical} & 14.8 & 7.4 & 5.7 & 4.0 & 7.3 \\
        \bottomrule
    \end{tabular}
    \vspace{-5pt}
    \caption{Performance of state-of-the-art 3DSS models that are pre-trained over SemanticKITTI and tested on validation set of SemanticSTF for individual weather conditions and jointly for \textit{all} weather conditions.}
    \label{tab:pretrained}
    \end{footnotesize}
\end{table}
}

\section{Conclusion and Outlook}
This paper presents SemanticSTF, a large-scale dataset and benchmark suite for semantic segmentation of LiDAR point clouds under adverse weather conditions. SemanticSTF provides high-quality point-level annotations for point clouds captured under adverse weather including dense fog, light fog, snow and rain. Extensive studies have been conducted to examine how state-of-the-art 3DSS methods perform over SemanticSTF, demonstrating its significance in directing future research on domain adaptive and domain generalizable 3DSS under all-weather conditions.

We also design PointDR, a domain randomization technique that aims to use normal-weather point clouds to train a domain generalizable 3DSS model that can work well over adverse-weather point clouds. PointDR consists of two novel designs including geometry style randomization and embedding aggregation which jointly learn perturbation-invariant representations that generalize well to various new point-cloud domains. Extensive experiments show that PointDR achieves superior point cloud segmentation performance as compared with the state-of-the-art.

\section*{Acknowledgement}
This study is funded BY the Ministry of Education Singapore, under the Tier-1 scheme with project number RG18/22. It is also supported under the RIE2020 Industry Alignment Fund – Industry Collaboration Projects (IAF-ICP) Funding Initiative, as well as cash and in-kind contribution from Singapore Telecommunications Limited (Singtel), through Singtel Cognitive and Artificial Intelligence Lab for Enterprises (SCALE@NTU).

%%%%%%%%% REFERENCES
{\small
\bibliographystyle{ieee_fullname}
\bibliography{egbib}
}

\clearpage

\appendix
\onecolumn

We provide more experiment details of domain adaptation and domain generalization in Section~\ref{DG} and Section~\ref{DA}, respectively, supervised learning on adverse conditions in Section~\ref{Sup3DSS} and additional details on SemanticSTF dataset in Section~\ref{SemanticsSTF}.

\section{Domain generalization}
\label{DG}

\subsection{Implementation details}
We provide the detailed training configurations for semantic segmentation of LiDAR point clouds that have been adopted as described in Sec. 5.1 of the submitted paper. Specifically, we implement the backbone model MinkowskiNet~\cite{choy20194d}  with the TorchSparse library~\cite{tang2022torchsparse}. 
For training, we use SGD optimizer. The learning rate, momentum and weight decays are set as $0.24, 0.9$, and $1.4e-4$, respectively. $\tau$ in Eq. 1 in the paper is set as 0.07~\cite{wu2018unsupervised,he2020momentum} and $\lambda_{ct}$ in Eq. 2 is set as $0.1$. The momentum coefficient $m$ is set at $0.99$. We train $50$ epochs with one NVIDIA 2080Ti with 11GB GPU memory and set the batch size at $4$.
The augmentations of training data in the source-domain are implemented as follows: For \textit{rotation}, LiDAR points are rotated with the range of  $[0,360\degree]$ along Z axis. For \textit{scale}, the coordinates of LiDAR points are randomly scaled within $[0.95,1.05]$. For \textit{drop-out}, we randomly drop-out 0-20\% points of input LiDAR scans with a probability of $0.5$.  As for \textit{noise perturbation}, $0-2,000$ random points are added into the 3D space of each LiDAR scan with a probability of $0.5$. When using \textit{flipping}, we randomly flip coordinates of LiDAR point clouds along x or y axis with a probability of $0.5$. As for \textit{jittering}, random coordinate shifts with a range of $[-0.05, 0.05]$ meters are added into LiDAR points with a probability of $0.5$.

In training the oracle model, we employ the SGD optimizer with the hyperparameters including initial learning rate at $0.1$, momentum at $0.9$, weight decay at $1.0e-4$, and dampening at $0.1$. We train the segmentation model with $500$ epochs using a single NVIDIA 2080Ti with 11GB GPU memory. The batch size is set as $4$. We use Poly learning rate policy with power$=0.9$. As for data augmentations, we follow~\cite{tang2020searching} and adoptes random rotation ($[-\pi, \pi]$) and scaling ($[0.95, 1.05]$); We also adopts PolarMix \cite{xiao2022polarmix} with following parameter settings: Rotation angles along the Z-axis, denoted as $\Omega$, are randomly scaled within normal distributions with a mean of $\mu=0$ and standard deviation of $\sigma=\frac{2}{3}\pi$. We keep the original instance classes for rotate-pasting in PolarMix. 

\subsection{Evaluation of individual adverse weather conditions}
We noticed that for certain individual adverse weather conditions, some class has no data captured in the validation set of SemanticSTF.   Specifically, there are no points of \textit{bicycle} and \textit{motorcycle} in the validation set of \textit{dense fog}; no points of \textit{bicyclist} and \textit{motorcyclist} in the validation set of \textit{snow}, and no \textit{bicycle} and \textit{motorcyclist} in the validation set of \textit{rain}. This is reasonable as the LiDAR data of SemanticSTF is collected in European countries including Germany, Sweden, Denmark, and Finland where motorcycles are not widely used for the reason of environmental protection. In addition, people usually do not ride bicycles or motorcycles in adverse weather conditions. As a result, classes \textit{motor}, \textit{motorcyclist}, and \textit{bicyclist} have extremely lower occurrence frequency, leading to an absence of these classes in the validation set of SemanticSTF under relevant weather conditions. Tables~\ref{tab:dg_stf_dense_fog}, \ref{tab:dg_stf_light_fog}, \ref{tab:dg_stf_rain}, and \ref{tab:dg_stf_snow} present corresponding class-level IoU performance for each adverse weather in Table 3 of the submitted paper.

\renewcommand\arraystretch{0.8}
\setlength{\tabcolsep}{1.9mm}{
\begin{table*}[t]
\centering
\begin{footnotesize}
\begin{tabular}{l|cccccccccccccccccccc}
 \toprule
  Methods & \rotatebox{90}{car} & \rotatebox{90}{bi.cle} & \rotatebox{90}{mt.cle} & \rotatebox{90}{truck} & \rotatebox{90}{oth-v.} & \rotatebox{90}{pers.} & \rotatebox{90}{bi.clst} & \rotatebox{90}{mt.clst} & \rotatebox{90}{road} & \rotatebox{90}{parki.} & \rotatebox{90}{sidew.} & \rotatebox{90}{oth-g.} & \rotatebox{90}{build.} & \rotatebox{90}{fence} & \rotatebox{90}{veget.} & \rotatebox{90}{trunk} & \rotatebox{90}{terra.} & \rotatebox{90}{pole} & \rotatebox{90}{traf.} \\
  \midrule
  \multicolumn{20}{c}{SemanticKITTI$\rightarrow$SemanticSTF(\textit{dense fog})}\\
  \midrule
  Baseline &  74.7 & - & - & 7.8 & 0.0 & 6.4 & 8.9 & 0.0 & 72.2 & 0.6 & 33.8 & 0.0 & 59.6 & 48.7 & 56.9 & 27.4 & 56.4 & 27.2 & 21.1\\
  Dropout~\cite{srivastava2014dropout} & 67.5 & - & - & 1.9 & 0.0 & 8.9 & 2.8 & 0.0 & 70.9 & 5.6 & 29.0 & 0.8 & 64.6 & 44.0 & 60.0 & 31.6 & 60.6 & 28.1 & 21.3 \\
  Perturbation & 68.6 & - & - & 8.8 & 0.0 & 6.0 & 0.0 & 0.0 & 66.6 & 14.8 & 24.3 & 0.1 & 52.2 & 43.5 & 60.1 & 19.4 & 54.1 & 16.3 & 11.5 \\
  PolarMix~\cite{xiao2022polarmix} & 52.3 & - & - & 17.2 & 0.0 & 3.6 & 0.0 & 19.3 & 75.2 & 0.0 & 28.7 & 0.6 & 62.4 & 49.5 & 60.5 & 29.0 & 55.4 & 20.8 & 30.7 \\
  MMD~\cite{li2018domain} & 75.5 & - & - & 0.3 & 0.0 & 4.2 & 0.0 & 0.0 & 75.4 & 11.2 & 33.6 & 0.5 & 64.8 & 51.7 & 64.7 & 26.1 & 62.3 & 23.0 & 23.0\\
  PCL~\cite{yao2022pcl} & 64.3 & - & - & 11.7 & 0.0 & 0.6 & 0.0 & 0.0 & 72.4 & 3.8 & 31.3 & 0.8 & 63.1 & 46.5 & 65.7 & 19.4 & 64.3 & 18.5 & 28.9\\
  \rowcolor{Gainsboro!60} PointDR (Ours) & 69.2 & - & - & 7.1 & 0.0 & 2.4 & 6.7 & 0.0 & 73.5 & 8.5 & 33.6 & 0.2 & 65.6 & 47.6 & 63.6 & 31.0 & 60.7 & 24.4 & 38.8\\
  \toprule
  \multicolumn{20}{c}{SynLiDAR$\rightarrow$SemanticSTF(\textit{dense fog})}\\
  \midrule
  Baseline & 21.6 & - & - & 6.4 & 0.0 & 3.7 & 2.9 & 18.9 & 25.7 & 0.0 & 7.7 & 1.0 & 41.2 & 22.5 & 52.3 & 15.4 & 55.5 & 9.3 & 2.4\\
  Dropout~\cite{srivastava2014dropout} & 12.7 & - & - & 7.7 & 0.0 & 1.9 & 0.4 & 2.5 & 38.3 & 0.1 & 10.2 & 0.3 & 37.3 & 21.8 & 57.4 & 13.1 & 44.5 & 10.1 & 1.0 \\
  Perturbation & 13.3 & - & - & 10.4 & 0.0 & 4.3 & 2.8 & 19.1 & 30.0 & 0.7 & 8.8 & 1.2 & 30.5 & 17.5 & 48.9 & 18.4 & 50.3 & 16.3 & 5.2\\
  PolarMix~\cite{xiao2022polarmix} &  15.8 & - & - & 10.6 & 0.0 & 1.5 & 1.7 & 3.5 & 27.7 & 0.0 & 9.9 & 0.3 & 46.2 & 28.9 & 59.2 & 13.5 & 49.5 & 4.4 & 1.7 \\
  MMD~\cite{li2018domain} & 26.5 & - & - & 12.7 & 0.0 & 2.7 & 4.0 & 22.3 & 30.6 & 0.0 & 9.4 & 0.0 & 31.6 & 21.7 & 52.6 & 13.9 & 54.3 & 8.9 & 2.5 \\
  PCL~\cite{yao2022pcl} & 22.9 & - & - & 20.1 & 0.0 & 2.2 & 6.2 & 28.3 & 29.0 & 0.0 & 9.2 & 2.6 & 37.9 & 22.9 & 54.5 & 11.4 & 45.9 & 8.5 & 1.1 \\
  \rowcolor{Gainsboro!60} PointDR (Ours) & 42.5 & - & - & 16.6 & 0.0 & 2.4 & 3.2 & 12.2 & 31.9 & 0.2 & 9.0 & 0.8 & 42.8 & 27.1 & 59.8 & 18.3 & 44.0 & 15.4 & 5.7 \\
    \bottomrule
    \end{tabular}
    \vspace{-10pt}
    \caption{Class-wise IoU on domain generalization with SemanticKITTI or SynLiDAR as the source and validation set of \textit{dense fog} in SemanticSTF as the target. ’-’ represents no samples captured in \textit{dense fog} in the validation set of SemanticSTF.}
    \label{tab:dg_stf_dense_fog}
  \end{footnotesize}
\end{table*}
}

\setlength{\tabcolsep}{1.8mm}{
\begin{table}[t]
\centering
\begin{footnotesize}
\begin{tabular}{l|ccccccccccccccccccccc}
 \toprule
  Methods & \rotatebox{90}{car} & \rotatebox{90}{bi.cle} & \rotatebox{90}{mt.cle} & \rotatebox{90}{truck} & \rotatebox{90}{oth-v.} & \rotatebox{90}{pers.} & \rotatebox{90}{bi.clst} & \rotatebox{90}{mt.clst} & \rotatebox{90}{road} & \rotatebox{90}{parki.} & \rotatebox{90}{sidew.} & \rotatebox{90}{oth-g.} & \rotatebox{90}{build.} & \rotatebox{90}{fence} & \rotatebox{90}{veget.} & \rotatebox{90}{trunk} & \rotatebox{90}{terra.} & \rotatebox{90}{pole} & \rotatebox{90}{traf.} \\
  \midrule
  \multicolumn{20}{c}{SemanticKITTI$\rightarrow$SemanticSTF(\textit{light fog})}\\
  \midrule
  Baseline &  60.0 & 0.0 & 0.0 & 1.3 & 10.9 & 12.3 & 0.0 & 0.0 & 68.6 & 4.5 & 36.0 & 0.0 & 61.5 & 53.1 & 55.6 & 38.0 & 44.7 & 29.2 & 18.2 \\
  Dropout~\cite{srivastava2014dropout} & 63.2 & 0.0 & 0.0 & 3.2 & 10.2 & 5.5 & 0.0 & 0.0 & 63.8 & 4.9 & 29.4 & 0.1 & 62.5 & 53.1 & 58.6 & 42.5 & 46.6 & 27.8 & 14.3 \\
  Perturbation & 76.6 & 0.0 & 0.0 & 38.2 & 0.0 & 21.9 & 0.0 & 0.0 & 66.6 & 8.8 & 34.6 & 0.1 & 62.4 & 56.1 & 63.2 & 25.3 & 46.2 & 22.4 & 6.5 \\
  PolarMix~\cite{xiao2022polarmix} & 42.6 & 0.2 & 0.0 & 29.4 & 3.3 & 17.0 & 0.0 & 0.2 & 69.8 & 0.7 & 33.1 & 0.1 & 56.2 & 56.3 & 54.9 & 24.7 & 44.8 & 24.1 & 16.6 \\
  MMD~\cite{li2018domain} & 63.6 & 0.0 & 0.0 & 0.1 & 13.3 & 25.9 & 0.0 & 0.0 & 73.9 & 5.6 & 42.8 & 0.1 & 64.1 & 55.3 & 61.9 & 36.6 & 50.7 & 29.2 & 9.9\\
  PCL~\cite{yao2022pcl} & 66.3 & 0.0 & 0.0 & 26.7 & 0.2 & 8.7 & 0.0 & 0.0 & 67.8 & 5.0 & 36.7 & 0.4 & 64.3 & 58.0 & 66.1 & 21.2 & 53.1 & 25.5 & 24.6\\
  \rowcolor{Gainsboro!60} PointDR (Ours) & 65.9 & 0.0 & 0.0 & 29.7 & 4.4 & 11.4 & 0.9 & 0.0 & 70.9 & 8.8 & 43.3 & 0.0 & 66.5 & 55.1 & 61.3 & 43.0 & 49.1 & 29.1 & 24.3\\
  \toprule
  \multicolumn{20}{c}{SynLiDAR$\rightarrow$SemanticSTF(\textit{light fog})}\\
  \midrule
  Baseline & 32.0 & 4.2 & 0.5 & 27.3 & 0.2 & 14.0 & 6.2 & 0.0 & 31.0 & 0.0 & 12.6 & 0.9 & 38.7 & 24.8 & 51.5 & 26.7 & 46.4 & 8.5 & 1.3\\
  Dropout~\cite{srivastava2014dropout} & 22.5 & 3.0 & 0.9 & 16.0 & 0.1 & 10.0 & 5.2 & 0.2 & 40.3 & 1.3 & 18.1 & 0.0 & 38.9 & 22.1 & 57.6 & 23.5 & 38.5 & 13.8 & 3.7 \\
  Perturbation & 31.1 & 1.9 & 1.6 & 21.5 & 0.0 & 12.5 & 2.6 & 0.0 & 33.2 & 1.6 & 14.3 & 1.1 & 34.3 & 20.1 & 48.7 & 29.8 & 42.0 & 16.7 & 4.5\\
  PolarMix~\cite{xiao2022polarmix} & 27.3 & 0.3 & 0.4 & 8.9 & 1.4 & 8.2 & 1.2 & 0.0 & 29.0 & 0.2 & 15.5 & 0.7 & 39.9 & 27.4 & 57.3 & 28.8 & 40.9 & 5.8 & 1.5 \\
  MMD~\cite{li2018domain} & 31.0 & 2.1 & 0.5 & 16.0 & 0.0 & 10.5 & 1.7 & 0.0 & 37.7 & 0.3 & 16.3 & 0.6 & 29.2 & 24.9 & 51.8 & 29.6 & 47.8 & 8.3 & 1.8 \\
  PCL~\cite{yao2022pcl} & 31.7 & 0.7 & 0.8 & 10.1 & 0.1 & 10.2 & 21.6 & 0.0 & 33.9 & 0.6 & 16.1 & 0.1 & 37.8 & 22.2 & 52.5 & 23.8 & 42.6 & 11.3 & 2.2 \\
  \rowcolor{Gainsboro!60} PointDR (Ours) & 44.7 & 1.7 & 1.0 & 33.9 & 0.3 & 12.9 & 4.7 & 0.0 & 36.0 & 0.9 & 15.8 & 0.7 & 44.4 & 30.3 & 60.0 & 28.3 & 42.4 & 15.1 & 5.7 \\
    \bottomrule
    \end{tabular}
    \vspace{-10pt}
    \caption{Class-wise IoU on domain generalization with SemanticKITTI or SynLiDAR as the source and validation set of \textit{light fog} in SemanticSTF as the target.}
    \label{tab:dg_stf_light_fog}
  \end{footnotesize}
\end{table}
}

\setlength{\tabcolsep}{1.9mm}{
\begin{table}[t]
\centering
\begin{footnotesize}
\begin{tabular}{l|ccccccccccccccccccccc}
 \toprule
  Methods & \rotatebox{90}{car} & \rotatebox{90}{bi.cle} & \rotatebox{90}{mt.cle} & \rotatebox{90}{truck} & \rotatebox{90}{oth-v.} & \rotatebox{90}{pers.} & \rotatebox{90}{bi.clst} & \rotatebox{90}{mt.clst} & \rotatebox{90}{road} & \rotatebox{90}{parki.} & \rotatebox{90}{sidew.} & \rotatebox{90}{oth-g.} & \rotatebox{90}{buid.} & \rotatebox{90}{fence} & \rotatebox{90}{veget.} & \rotatebox{90}{trunk} & \rotatebox{90}{terra.} & \rotatebox{90}{pole} & \rotatebox{90}{traf.} \\
  \midrule
  \multicolumn{20}{c}{SemanticKITTI$\rightarrow$SemanticSTF(\textit{rain})}\\
  \midrule
  Baseline &  72.4 & 0.0 & - & 0.0 & 16.3 & 6.9 & 0.0 & - & 71.6 & 12.7 & 58.1 & 0.0 & 70.0 & 33.0 & 51.8 & 9.9 & 24.2 & 33.3 & 22.9 \\
  Dropout~\cite{srivastava2014dropout} & 81.3 & 0.0 & - & 0.0 & 21.2 & 5.6 & 0.0 & - & 62.2 & 11.8 & 44.8 & 0.6 & 76.8 & 44.7 & 56.0 & 16.3 & 23.3 & 32.8 & 22.2 \\
  Perturbation & 83.9 & 0.0 & - & 2.4 & 0.0 & 20.9 & 0.0 & - & 73.2 & 12.6 & 54.7 & 7.0 & 71.7 & 43.2 & 58.3 & 5.9 & 29.4 & 29.4 & 16.9 \\
  PolarMix~\cite{xiao2022polarmix} & 56.7 & 4.0 & - & 9.1 & 1.5 & 29.8 & 0.0 & - & 68.2 & 10.9 & 50.2 & 0.5 & 73.2 & 47.2 & 48.3 & 17.8 & 22.3 & 32.3 & 14.1 \\
  MMD~\cite{li2018domain} & 83.9 & 0.0 & - & 0.0 & 8.9 & 31.6 & 0.0 & - & 77.9 & 17.9 & 60.2 & 0.3 & 69.6 & 39.3 & 58.4 & 14.1 & 32.5 & 34.0 & 30.0 \\
  PCL~\cite{yao2022pcl} & 84.2 & 0.0 & - & 0.0 & 0.1 & 4.3 & 0.0 & - & 68.1 & 10.9 & 55.5 & 4.6 & 74.7 & 43.9 & 59.6 & 5.8 & 27.3 & 34.2 & 38.8\\
  \rowcolor{Gainsboro!60} PointDR (Ours) & 78.0 & 0.0 & - & 0.0 & 13.8 & 20.0 & 0.0 & - & 72.1 & 14.7 & 60.0 & 1.2 & 76.1 & 36.9 & 58.0 & 18.3 & 24.7 & 36.1 & 32.5\\
  \toprule
  \multicolumn{20}{c}{SynLiDAR$\rightarrow$SemanticSTF(\textit{rain})}\\
  \midrule
  Baseline & 45.8 & 4.5 & - & 6.8 & 0.4 & 38.9 & 0.0 & - & 32.0 & 0.0 & 24.3 & 0.0 & 43.0 & 8.0 & 33.8 & 11.3 & 23.9 & 11.5 & 7.7\\
  Dropout~\cite{srivastava2014dropout} & 47.0 & 7.6 & - & 7.7 & 0.0 & 34.0 & 0.0 & - & 47.3 & 6.9 & 34.6 & 0.0 & 39.8 & 11.5 & 37.5 & 13.8 & 29.6 & 21.6 & 8.6 \\
  Perturbation & 57.5 & 5.3 & - & 18.2 & 0.0 & 36.3 & 0.1 & - & 37.1 & 1.5 & 26.9 & 0.3 & 34.9 & 10.4 & 32.6 & 12.2 & 20.5 & 23.2 & 10.4\\
  PolarMix~\cite{xiao2022polarmix} &  59.6 & 1.5 & - & 6.0 & 5.2 & 24.6 & 1.0 & - & 31.4 & 0.1 & 30.4 & 0.0 & 55.5 & 12.2 & 44.6 & 13.1 & 25.0 & 11.0 & 4.7 \\
  MMD~\cite{li2018domain} & 49.5 & 4.8 & - & 20.0 & 4.7 & 37.6 & 0.0 & - & 43.7 & 0.0 & 32.4 & 0.0 & 42.1 & 11.3 & 34.4 & 12.3 & 25.1 & 13.4 & 8.1 \\
  PCL~\cite{yao2022pcl} & 51.3 & 0.9 & - & 4.3 & 2.1 & 35.6 & 0.0 & - & 41.4 & 0.0 & 32.0 & 0.0 & 54.8 & 9.7 & 37.1 & 11.4 & 24.2 & 16.6 & 6.3 \\
  \rowcolor{Gainsboro!60} PointDR (Ours) & 42.2 & 3.3 & - & 21.9 & 0.0 & 30.4 & 1.7 & - & 35.8 & 3.2 & 31.9 & 0.0 & 54.0 & 14.4 & 40.7 & 12.5 & 31.9 & 23.6 & 11.8 \\
    \bottomrule
    \end{tabular}
    \vspace{-10pt}
    \caption{Class-wise IoU on domain generalization with SemanticKITTI or SynLiDAR as the source and validation set of \textit{rain} in SemanticSTF as the target. ’-’ represents no samples captured in \textit{rain} in the validation set of SemanticSTF.}
    \label{tab:dg_stf_rain}
  \end{footnotesize}
\end{table}

\setlength{\tabcolsep}{1.9mm}{
\begin{table*}[t]
\centering
\begin{footnotesize}
\begin{tabular}{l|ccccccccccccccccccccc}
 \toprule
  Methods & \rotatebox{90}{car} & \rotatebox{90}{bi.cle} & \rotatebox{90}{mt.cle} & \rotatebox{90}{truck} & \rotatebox{90}{oth-v.} & \rotatebox{90}{pers.} & \rotatebox{90}{bi.clst} & \rotatebox{90}{mt.clst} & \rotatebox{90}{road} & \rotatebox{90}{parki.} & \rotatebox{90}{sidew.} & \rotatebox{90}{oth-g.} & \rotatebox{90}{build.} & \rotatebox{90}{fence} & \rotatebox{90}{veget.} & \rotatebox{90}{trunk} & \rotatebox{90}{terra.} & \rotatebox{90}{pole} & \rotatebox{90}{traf.} \\
  \midrule
  \multicolumn{20}{c}{SemanticKITTI$\rightarrow$SemanticSTF(\textit{snow})}\\
  \midrule
  Baseline & 49.5 & 0.0 & 0.3 & 0.5 & 11.6 & 10.8 & - & - & 42.1 & 14.9 & 23.9 & 0.0 & 71.5 & 26.7 & 29.3 & 24.0 & 17.8 & 30.8 & 10.1 \\
  Dropout~\cite{srivastava2014dropout} & 58.5 & 0.0 & 30.5 & 5.4 & 13.2 & 5.2 & - & - & 41.9 & 18.0 & 20.4 & 2.5 & 76.4 & 30.5 & 31.8 & 32.7 & 19.8 & 28.2 & 7.0 \\
  Perturbation & 73.6 & 0.0 & 0.0 & 5.5 & 1.1 & 19.8 & - & - & 45.7 & 10.9 & 34.4 & 0.1 & 80.6 & 32.8 & 45.2 & 12.8 & 20.0 & 24.4 & 9.5\\
  PolarMix~\cite{xiao2022polarmix} &  66.5 & 3.4 & 9.3 & 3.5 & 5.8 & 32.4 & - & - & 55.3 & 3.6 & 30.1 & 0.1 & 77.8 & 36.1 & 34.2 & 12.6 & 25.1 & 29.8 & 10.1 \\
  MMD~\cite{li2018domain} & 59.4 & 0.0 & 4.7 & 0.0 & 14.7 & 30.5 & - & - & 50.8 & 16.9 & 32.8 & 0.2 & 68.4 & 24.4 & 36.6 & 24.1 & 24.1 & 30.0 & 11.4\\
  PCL~\cite{yao2022pcl} & 64.0 & 0.0 & 0.0 & 8.2 & 0.7 & 9.2 & - & - & 38.9 & 15.2 & 31.6 & 2.3 & 79.6 & 35.1 & 41.3 & 11.2 & 23.1 & 30.1 & 26.8\\
  \rowcolor{Gainsboro!60} PointDR (Ours) & 66.2 & 0.0 & 10.4 & 0.0 & 16.7 & 21.3 & - & - & 43.0 & 15.2 & 33.0 & 1.7 & 76.8 & 30.3 & 36.1 & 27.6 & 22.2 & 30.0 & 14.1\\
  \toprule
  \multicolumn{20}{c}{SynLiDAR$\rightarrow$SemanticSTF(\textit{snow})}\\
  \midrule
  Baseline & 24.6 & 2.7 & 1.5 & 2.4 & 0.0 & 32.2 & - & - & 12.9 & 0.4 & 18.3 & 0.0 & 33.3 & 13.8 & 15.7 & 14.9 & 18.1 & 10.1 & 1.9\\
  Dropout~\cite{srivastava2014dropout} & 35.9 & 2.8 & 3.7 & 3.0 & 0.0 & 21.9 & - & - & 20.9 & 10.0 & 22.8 & 0.0 & 33.2 & 14.8 & 17.1 & 16.8 & 16.5 & 15.7 & 2.6 \\
  Perturbation & 27.1 & 2.4 & 6.8 & 6.8 & 0.2 & 31.0 & - & - & 15.4 & 4.8 & 19.7 & 0.0 & 26.3 & 12.4 & 14.0 & 22.0 & 16.4 & 19.0 & 4.1 \\
  PolarMix~\cite{xiao2022polarmix} & 53.4 & 2.3 & 4.1 & 6.0 & 1.2 & 27.9 & - & - & 11.7 & 1.9 & 21.5 & 0.3 & 45.2 & 20.8 & 21.7 & 18.8 & 16.5 & 10.5 & 1.7 \\
  MMD~\cite{li2018domain} & 20.8 & 2.7 & 6.0 & 4.8 & 0.2 & 31.3 & - & - & 20.1 & 0.5 & 21.0 & 0.1 & 29.6 & 12.2 & 15.0 & 16.6 & 21.8 & 11.3 & 2.4 \\
  PCL~\cite{yao2022pcl} & 30.7 & 1.1 & 4.4 & 6.2 & 0.3 & 34.6 & - & - & 19.1 & 1.7 & 22.0 & 0.3 & 37.8 & 12.6 & 16.4 & 14.2 & 19.9 & 14.7 & 3.0 \\
  \rowcolor{Gainsboro!60} PointDR (Ours) & 34.2 & 4.0 & 7.4 & 7.5 & 0.1 & 36.2 & - & - & 13.8 & 12.0 & 22.7 & 0.0 & 48.8 & 19.9 & 19.9 & 18.9 & 17.0 & 20.7 & 3.4 \\
    \bottomrule
    \end{tabular}
    \vspace{-10pt}
    \caption{Class-wise IoU on domain generalization with SemanticKITTI or SynLiDAR as the source and validation set of \textit{snow} in SemanticSTF as the target. ’-’ represents no samples captured in \textit{snow} in the validation set of SemanticSTF.}
    \label{tab:dg_stf_snow}
  \end{footnotesize}
\end{table*}
}

\subsection{Ablation study}

\noindent\textbf{Data augmentation}. We study how data augmentation techniques affect generalized semantic segmentation of point clouds (3DSS) and compare them with the proposed PointDR. As Table~\ref{tab:aug} shows, we report seven models over the benchmark “SemanticKITTI$\rightarrow$SemanticSTF”: 1) The \textit{Baseline} is a source-only model that is trained by using the training data of SemanticKITTI; 2) The \textit{drop-out, noise perturbation, flipping}, and \textit{jittering} are segmentation models with different augmentation techniques over input data; \textit{All} is the model that combines of all these augmentation techniques; 3) Our proposed \textit{PointDR}. 
We can see that implementing each of these augmentation techniques improves the generalization capability of the segmentation model clearly and consistently. However, the combination of them all did not yield the best segmentation performance, largely because the combination brings too many distortions to the input point clouds. On the contrary, the proposed PointDR achieves the best segmentation performance, indicating its superior ability to learn universal representations for all-weather 3DSS.

\setlength{\tabcolsep}{3.0mm}{
\begin{table}[h]
    \centering
    \begin{footnotesize}
    \begin{tabular}{l|c|ccccc|c}
    \toprule
        Method & Baseline & drop-out & perturbation & flipping & jittering & All & PointDR \\
    \midrule
        mIoU & 24.4 & 25.7 & 25.9 & 25.2 & 26.9 & 26.1& 28.6\\
    \bottomrule
    \end{tabular}
    \vspace{-10pt}
    \caption{Comparison of data augmentation techniques and the proposed PointDR.  PointDR performs clearly the best over domain generalized segmentation task SemanticKITTI$\rightarrow$SemanticSTF.}
    \label{tab:aug}
    \end{footnotesize}
\end{table}
}

\noindent\textbf{Parameter analysis}. We examine the parameter $\lambda_{cl}$ in Eq. 2 in the paper that balances the cross entropy loss and the contrastive loss. As Table~\ref{tab:ablation_weight1} shows, optimizing the proposed contrastive loss is able to improve segmentation performance consistently while different $\lambda_{cl}$ produce quite different mIoUs. The best mIoU is obtained when $\lambda_{cl}=0.10$.

\setlength{\tabcolsep}{3.0mm}{
\begin{table}[h]
    \centering
    \begin{footnotesize}
    \begin{tabular}{c|ccccc}
    \toprule
        $\lambda_{cl}$ & 0.0 & 0.05 & 0.10 & 0.15 & 1.0\\
    \midrule
        mIoU  & 24.4 & 28.2 & 28.6 & 27.3 & 25.1 \\
    \bottomrule
    \end{tabular}
    \vspace{-10pt}
    \caption{Performance of PointDR models with different contrastive loss weight $\lambda_{cl}$ in Eq. 2 in the paper.}
    \label{tab:ablation_weight1}
    \end{footnotesize}
\end{table}
}

Table~\ref{tab:ablation_weight2} below shows segmentation performance with different momentum values ($m$) used for updating the memory bank $\mathcal{B}$. It performs reasonably well when $m$ is $0.98$ or $0.99$, showing that a slowly progressing memory bank is beneficial. However, when $m$ is too large (at $0.999$), the memory bank updates too slowly to capture the latest and representative feature embeddings, which fails to serve as the class-wise proxy and ultimately leads to a clear segmentation performance drop.

\setlength{\tabcolsep}{3.0mm}{
\begin{table}[h]
    \centering
    \begin{footnotesize}
    \begin{tabular}{c|ccc}
    \toprule
        $m$ & 0.98 & 0.99 & 0.999 \\
    \midrule
        mIoU  & 28.1 & 28.6 & 26.1\\
    \bottomrule
    \end{tabular}
    \vspace{-10pt}
    \caption{ Performance of PointDR models with different momentum updated weight $m$ for the memory bank $\mathcal{B}$.}
    \label{tab:ablation_weight2}
    \end{footnotesize}
\end{table}
}

\section{Domain adaptation}
\label{DA}

\subsection{Implementation details}
In Tables 4 and 5 of the submitted paper, we examine state-of-the-art UDA methods over the proposed \textit{normal-to-adverse} UDA scenario. Specifically, we selected typical UDA methods from the popular \textit{synthetic-to-real} UDA benchmark~\cite{xiao2022transfer,saltori2022cosmix} as the baseline methods as described in Section 5.2 of the paper. We adopt MinkowskiNet~\cite{choy20194d} as the segmentation model as in \textit{synthetic-to-real} UDA. When implementing ADDA~\cite{tzeng2017adda}, entropy minimization~\cite{vu2019advent}, and self-training~\cite{zou2019confidence}, we follow the same implementation and training configurations as the \textit{synthetic-to-real} UDA~\cite{xiao2022transfer} and leverage TorchSparse library~\cite{tang2022torchsparse}] (with version 1.1.0) based on PyTorch~\cite{paszke2019pytorch} library. While for CoSMix~\cite{saltori2022cosmix}, we use the officially released codes based on MinkowskiEngine with default training parameters for the adaptation. We report mIoU of the covered classes for individual adverse weather conditions in Table 5.

\subsection{Detailed class-level results} 
In Tables~\ref{tab:uda_kitti2stf_densefog}, \ref{tab:uda_kitti2stf_lightfog}, \ref{tab:uda_kitti2stf_rain}, and \ref{tab:uda_kitti2stf_snow} below, we present the class-level IoU performance for the UDA methods that are examined in the setting of adaptation to individual conditions in Table 5 of the paper.

\setlength{\tabcolsep}{1.8mm}{
\begin{table*}[!h]
\centering
\begin{footnotesize}
\begin{tabular}{l|cccccccccccccccccccc}
    \toprule
    Methods & \rotatebox{90}{car} & \rotatebox{90}{bi.cle} & \rotatebox{90}{mt.cle} & \rotatebox{90}{truck} & \rotatebox{90}{oth-v.} & \rotatebox{90}{pers.} & \rotatebox{90}{bi.clst} & \rotatebox{90}{mt.clst} & \rotatebox{90}{road} & \rotatebox{90}{parki.} & \rotatebox{90}{sidew.} & \rotatebox{90}{oth-g.} & \rotatebox{90}{build.} & \rotatebox{90}{fence} & \rotatebox{90}{veget.} & \rotatebox{90}{trunk} & \rotatebox{90}{terra.} & \rotatebox{90}{pole} & \rotatebox{90}{traf.}\\
    \midrule
    Source-only &  56.4 & - & - & 10.1 & 0.0 & 0.6 & 15.4 & 0.0 & 68.0 & 0.6 & 22.8 & 0.0 & 63.6 & 36.6 & 62.8 & 29.4 & 53.5 & 17.7 & 19.5\\
    ADDA~\cite{tzeng2017adda} & 63.4 & - & - & 14.3 & 0.0 & 2.1 & 8.0 & 38.7 & 68.0 & 0.1 & 25.6 & 0.0 & 60.6 & 45.4 & 64.8 & 30.4 & 52.6 & 20.4 & 41.9\\
    Ent-Min~\cite{vu2019advent} & 68.0 & - & - & 4.9 & 0.0 & 1.9 & 7.6 & 0.0 & 74.8 & 0.0 & 39.4 & 0.0 & 68.8 & 50.5 & 61.0 & 28.3 & 63.3 & 22.7 & 43.2\\
    Self-training~\cite{zou2019confidence} & 68.2 & - & - & 24.4 & 0.0 & 5.4 & 4.8 & 0.0 & 70.9 & 0.3 & 31.3 & 0.0 & 65.9 & 46.7 & 59.2 & 31.6 & 55.4 & 22.5 & 43.7 \\
    CoSMix~\cite{saltori2022cosmix} & 76.5 & - & - & 27.0 & 0.0 & 4.7 & 0.0 & 0.0 & 74.2 & 0.5 & 29.9 & 1.8 & 62.1 & 48.0 & 62.6 & 37.3 & 59.6 & 23.4 & 28.8\\
    \bottomrule
    \end{tabular}
    \vspace{-10pt}
    \caption{Comparison of state-of-the-art domain adaptation methods on SemanticKITTI$\rightarrow$SemanticSTF adaptation for \textit{dense fog}. ’-’ represents no samples captured in \textit{dense fog} in the validation set of SemanticSTF.
    }
    \label{tab:uda_kitti2stf_densefog}
    \end{footnotesize}
    \vspace{-7pt}
\end{table*}
}

\setlength{\tabcolsep}{1.7mm}{
\begin{table*}[!h]
\centering
\begin{footnotesize}
\begin{tabular}{l|cccccccccccccccccccc}
    \toprule
    Methods & \rotatebox{90}{car} & \rotatebox{90}{bi.cle} & \rotatebox{90}{mt.cle} & \rotatebox{90}{truck} & \rotatebox{90}{oth-v.} & \rotatebox{90}{pers.} & \rotatebox{90}{bi.clst} & \rotatebox{90}{mt.clst} & \rotatebox{90}{road} & \rotatebox{90}{parki.} & \rotatebox{90}{sidew.} & \rotatebox{90}{oth-g.} & \rotatebox{90}{build.} & \rotatebox{90}{fence} & \rotatebox{90}{veget.} & \rotatebox{90}{trunk} & \rotatebox{90}{terra.} & \rotatebox{90}{pole} & \rotatebox{90}{traf.}\\
    \midrule
    Source-only &  55.1 & 0.0 & 0.0 & 16.3 & 4.3 & 0.7 & 0.8 & 0.0 & 68.3 & 5.3 & 33.2 & 0.0 & 66.0 & 44.1 & 62.0 & 40.3 & 48.2 & 23.4 & 10.3\\
    ADDA~\cite{tzeng2017adda} & 61.4 & 0.0 & 0.0 & 40.0 & 10.8 & 1.4 & 2.3 & 0.0 & 69.4 & 2.5 & 36.3 & 0.0 & 62.0 & 52.0 & 60.4 & 43.2 & 48.9 & 22.7 & 16.9\\
    Ent-Min~\cite{vu2019advent} & 67.1 & 0.0 & 0.0 & 46.7 & 12.0 & 0.0 & 0.0 & 0.0 & 73.4 & 0.2 & 38.8 & 0.0 & 67.1 & 56.6 & 56.7 & 38.2 & 46.8 & 25.0 & 15.7\\
    Self-training~\cite{zou2019confidence} & 69.3 & 0.0 & 0.0 & 47.5 & 19.4 & 0.9 & 0.1 & 0.0 & 73.2 & 0.8 & 40.7 & 0.0 & 67.4 & 56.5 & 58.5 & 41.3 & 47.1 & 26.6 & 19.8 \\
    CoSMix~\cite{saltori2022cosmix} & 74.9 & 0.4 & 1.3 & 19.3 & 1.6 & 26.1 & 0.0 & 0.0 & 70.3 & 10.0 & 35.0 & 1.1 & 67.1 & 54.7 & 64.1 & 46.4 & 49.4 & 25.4 & 28.7\\
    \bottomrule
    \end{tabular}
    \vspace{-10pt}
    \caption{Comparison of state-of-the-art domain adaptation methods on SemanticKITTI$\rightarrow$SemanticSTF adaptation for \textit{light fog}.}
    \label{tab:uda_kitti2stf_lightfog}
    \end{footnotesize}
    \vspace{-7pt}
\end{table*}
}

\setlength{\tabcolsep}{1.8mm}{
\begin{table*}[!h]
\centering
\begin{footnotesize}
\begin{tabular}{l|cccccccccccccccccccc}
    \toprule
    Methods & \rotatebox{90}{car} & \rotatebox{90}{bi.cle} & \rotatebox{90}{mt.cle} & \rotatebox{90}{truck} & \rotatebox{90}{oth-v.} & \rotatebox{90}{pers.} & \rotatebox{90}{bi.clst} & \rotatebox{90}{mt.clst} & \rotatebox{90}{road} & \rotatebox{90}{parki.} & \rotatebox{90}{sidew.} & \rotatebox{90}{oth-g.} & \rotatebox{90}{build.} & \rotatebox{90}{fence} & \rotatebox{90}{veget.} & \rotatebox{90}{trunk} & \rotatebox{90}{terra.} & \rotatebox{90}{pole} & \rotatebox{90}{traf.}\\
    \midrule
    Source-only &  69.4 & 0.0 & - & 0.1 & 0.1 & 12.1 & 0.0 & - & 72.9 & 9.7 & 54.5 & 0.0 & 73.7 & 31.2 & 55.2 & 16.2 & 21.4 & 33.9 & 18.8\\
    ADDA~\cite{tzeng2017adda} & 71.8 & 0.0 & - & 0.1 & 0.7 & 3.8 & 0.0 & - & 71.9 & 9.2 & 51.5 & 0.0 & 67.8 & 35.6 & 53.6 & 17.8 & 25.7 & 32.0 & 24.2\\
    Ent-Min~\cite{vu2019advent} & 78.4 & 0.0 & - & 0.4 & 2.9 & 0.1 & 0.0 & - & 80.3 & 10.1 & 57.9 & 0.0 & 78.0 & 47.1 & 53.8 & 13.0 & 24.1 & 35.8 & 33.8\\
    Self-training~\cite{zou2019confidence} & 69.4 & 0.0 & - & 0.1 & 0.1 & 12.1 & 0.0 & - & 72.9 & 9.7 & 54.5 & 0.0 & 73.7 & 31.2 & 55.2 & 16.2 & 21.4 & 33.9 & 18.8 \\
    CoSMix~\cite{saltori2022cosmix} & 83.6 & 0.1 & - & 2.1 & 11.8 & 47.9 & 0.0 & - & 64.7 & 10.9 & 51.1 & 2.5 & 72.6 & 47.2 & 59.8 & 25.7 & 20.9 & 27.2 & 35.2\\
    \bottomrule
    \end{tabular}
    \vspace{-10pt}
    \caption{Comparison of state-of-the-art domain adaptation methods on SemanticKITTI$\rightarrow$SemanticSTF adaptation for \textit{rain}. ’-’ represents no samples captured in \textit{rain} in the validation set of SemanticSTF.
    }
    \label{tab:uda_kitti2stf_rain}
    \end{footnotesize}
    \vspace{-7pt}
\end{table*}
}

\setlength{\tabcolsep}{1.8mm}{
\begin{table*}[!h]
\centering
\begin{footnotesize}
\begin{tabular}{l|cccccccccccccccccccc}
    \toprule
    Methods & \rotatebox{90}{car} & \rotatebox{90}{bi.cle} & \rotatebox{90}{mt.cle} & \rotatebox{90}{truck} & \rotatebox{90}{oth-v.} & \rotatebox{90}{pers.} & \rotatebox{90}{bi.clst} & \rotatebox{90}{mt.clst} & \rotatebox{90}{road} & \rotatebox{90}{parki.} & \rotatebox{90}{sidew.} & \rotatebox{90}{oth-g.} & \rotatebox{90}{build.} & \rotatebox{90}{fence} & \rotatebox{90}{veget.} & \rotatebox{90}{trunk} & \rotatebox{90}{terra.} & \rotatebox{90}{pole} & \rotatebox{90}{traf.}\\
    \midrule
    Source-only &  70.7 & 0.0 & 0.0 & 15.4 & 1.6 & 5.1 & - & - & 49.8 & 8.9 & 36.6 & 0.0 & 67.1 & 26.3 & 30.7 & 28.1 & 22.1 & 26.8 & 9.6\\
    ADDA~\cite{tzeng2017adda} & 69.3 & 0.0 & 0.0 & 14.0 & 0.8 & 2.9 & - & - & 55.3 & 1.3 & 35.7 & 0.0 & 67.2 & 26.7 & 37.5 & 30.1 & 21.2 & 25.4 & 11.0\\
    Ent-Min~\cite{vu2019advent} & 73.8 & 0.0 & 15.4 & 19.8 & 1.4 & 2.9 & - & - & 53.6 & 1.6 & 32.9 & 0.0 & 73.4 & 28.5 & 34.1 & 28.8 & 21.7 & 26.6 & 8.8\\
    Self-training~\cite{zou2019confidence} & 73.9 & 0.0 & 6.1 & 16.9 & 5.2 & 7.7 & - & - & 53.9 & 6.2 & 34.3 & 0.0 & 69.3 & 27.7 & 33.7 & 29.8 & 19.5 & 26.9 & 16.0 \\
    CoSMix~\cite{saltori2022cosmix} & 79.2 & 1.3 & 0.0 & 0.6 & 14.2 & 38.9 & - & - & 70.1 & 15.1 & 54.1 & 6.3 & 74.6 & 44.1 & 58.3 & 20.5 & 20.4 & 26.9 & 35.6 \\
    \bottomrule
    \end{tabular}
    \vspace{-10pt}
    \caption{Comparison of state-of-the-art domain adaptation methods on SemanticKITTI$\rightarrow$SemanticSTF adaptation for \textit{snow}. ’-’ represents no samples captured in \textit{snow} in the validation set of SemanticSTF.
    }
    \label{tab:uda_kitti2stf_snow}
    \end{footnotesize}
\end{table*}
}

\setlength{\tabcolsep}{1.0mm}{
\begin{table*}[!h]%[t]
\centering
\begin{footnotesize}
\begin{tabular}{r|cccccccccccccccccccc|c}
 \toprule
  Methods & \rotatebox{90}{car} & \rotatebox{90}{bi.cle} & \rotatebox{90}{mt.cle} & \rotatebox{90}{truck} & \rotatebox{90}{oth-v.} & \rotatebox{90}{pers.} & \rotatebox{90}{bi.clst} & \rotatebox{90}{mt.clst} & \rotatebox{90}{road} & \rotatebox{90}{parki.} & \rotatebox{90}{sidew.} & \rotatebox{90}{oth-g.} & \rotatebox{90}{build.} & \rotatebox{90}{fence} & \rotatebox{90}{veget.} & \rotatebox{90}{trunk} & \rotatebox{90}{terra.} & \rotatebox{90}{pole} & \rotatebox{90}{traf.} &  \rotatebox{90}{invalid} & mIoU\\
 \midrule
  RandLA-Net~\cite{hu2020randla} & 75.2 & 0.0 & 0.0 & 25.8 & 0.0 & 47.3 & 0.0 & 0.0 & 73.3 & 7.8 & 48.7 & 57.5 & 68.2 & 48.3 & 61.5 & 27.3 & 49.5 & 39.7 & 27.5 & 56.5 & 35.7 \\
  SalsaNext~\cite{cortinhal2020salsanext} & 77.3 & 31.2 & 0.0 & 47.5 & 30.5 & 64.2 & 26.6 & 5.0 & 76.3 & 18.2 & 55.2 & 64.9 & 79.2 & 50.4 & 56.8 & 27.8 & 55.8 & 36.8 & 36.7 & 62.2 & 45.1 \\
  MinkowskiNet~\cite{choy20194d} & 87.4 & 42.5 & 0.0 & 51.2 & 40.3 & 73.6 & 29.1 & 0.0 & 79.5 & 15.0 & 57.7 & 63.4 & 78.6 & 56.8 & 64.4 & 40.4 & 53.3 & 47.6 & 47.6 & 67.7 & 49.8\\
  SPVCNN~\cite{tang2020searching} & 87.1 & 45.5 & 0.0 & 53.1 & 42.7 & 74.1 & 21.9 & 0.0 & 78.9 & 16.3 & 57.9 & 57.0 & 78.6 & 56.5 & 65.6 & 40.9 & 50.3 & 49.4 & 45.9 & 66.4 & 49.4\\ 
  Cylinder3D~\cite{zhu2021cylindrical} & 77.7 & 31.7 & 2.7 & 43.4 & 23.8 & 67.8 & 18.4 & 0.0 & 78.5 & 10.0 & 51.8 & 48.7 & 81.2 & 56.0 & 63.4 & 38.3 & 52.1 & 48.0 & 43.0 & 63.9  & 45.0 \\
  \bottomrule
\end{tabular}
\caption{Comparison of state-of-the-art 3DSS methods (trained in a supervised manner) over the test set of SemanticSTF.}
\label{tab:sup_stf}
\end{footnotesize}
\end{table*}
}

\section{Supervised learning on adverse conditions}
\label{Sup3DSS}

We use SemanticSTF to train five state-of-the-art 3DSS models in a supervised manner and report their segmentation performance in Table~\ref{tab:sup_stf}. 
Specifically, we use their officially released codes and default training configurations for model training. 
We can see that these state-of-the-art models achieve much lower segmentation performance over SemanticSTF as compared with their performance over SemanticKITTI. The results indicate that SemanticSTF is a more challenging benchmark for supervised methods due to the diverse data distribution and hard geometric domains. In addition, comparing Table~\ref{tab:sup_stf} and Table 6 of the submitted paper, we notice that the rankings of the supervised and the pre-trained 3DSS models are not well aligned, indicating that the ability of supervised representation learning may not be highly correlated with the generalization ability.
We also notice that the state-of-the-art network Cylinder3D~\cite{zhu2021cylindrical} achieves much lower segmentation performance over SemanticSTF as compared with its performance over SemanticKITTI. This could be due to two major factors: 1) The design of Cylinder3D is sensitive to complicated and noisy geometries of point clouds as introduced by various adverse weather conditions; 2) Cylinder3D is sensitive to training parameters and the default training configurations for SemanticKITTI does not work well for SemanticSTF. The results further demonstrate the importance of studying universal 3DSS as well as the value of the proposed SemanticSTF dataset  in steering the future endeavour along this very meaningful
research direction.

\section{Additional Details on SemanticSTF Dataset}\label{SemanticsSTF}

\subsection{Annotation}
In this section, we explain the implementation of our
point cloud labeling in more detail.  We leveraged a professional labeling program that has multiple annotating tools such as a brush, a polygon tool, a bounding volume tool, as well as different filtering methods for hiding labeled points or selected labels. Corresponding 2D images are displayed to assist labelling. The program also supports cross-checking and correction as illustrated in the main paper. Fig.~\ref{fig:labelling_tool} shows the interface of our point cloud annotation program.

\begin{figure}[!h]
    \centering
    \includegraphics[width=0.7\linewidth]{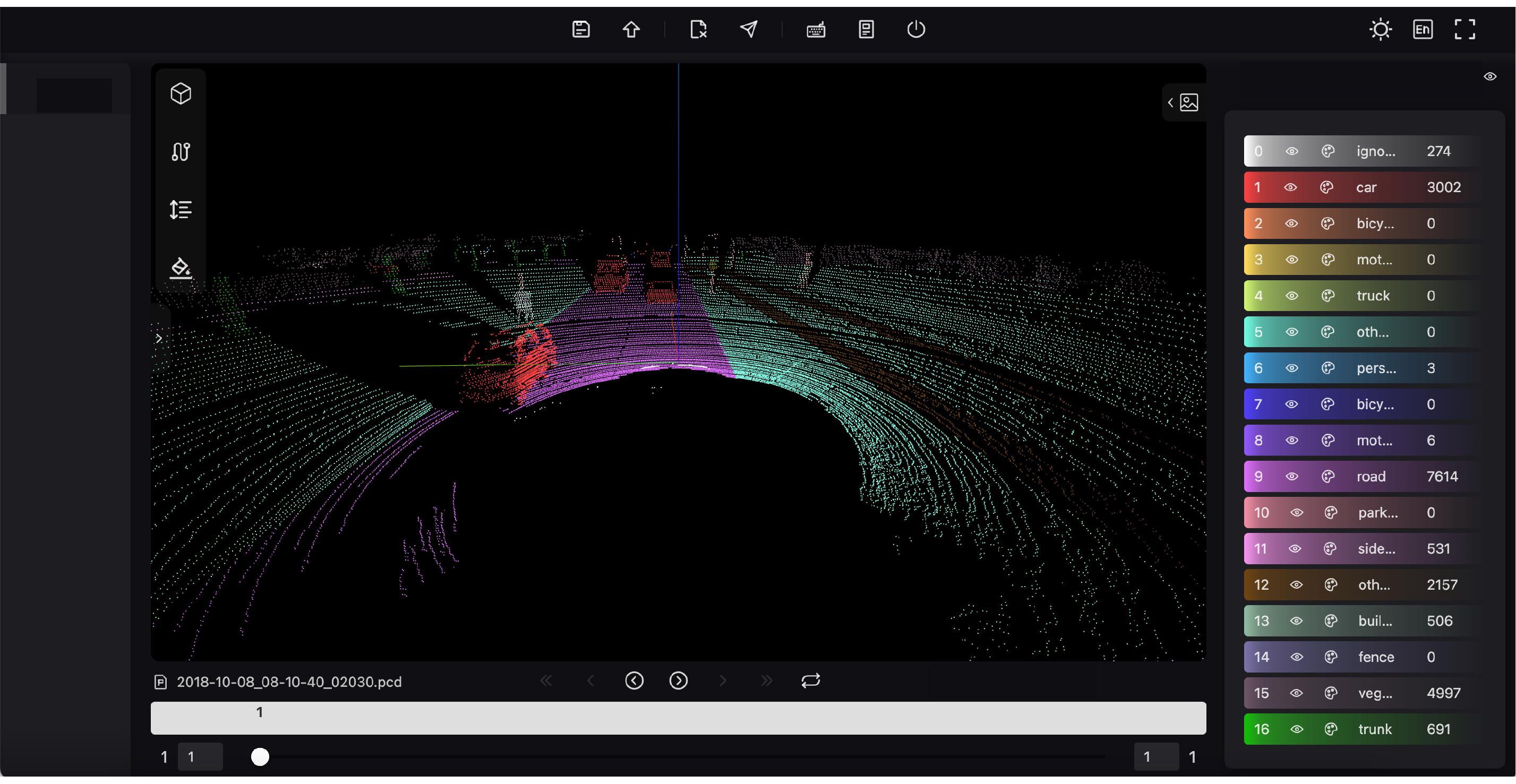}
    \vspace{-10pt}
    \caption{The interface of point cloud labeling program for annotating SemanticSTF.}
    \label{fig:labelling_tool}
\end{figure}

\subsection{Semantic class definition}

In the process of labeling such challenging data,
we had to decide which classes we wanted to annotate
at some point in time. In general, we followed the class
definitions of the SemanticKITTI dataset~\cite{behley2019semantickitti} and ACDC~\cite{sakaridis2021acdc} dataset, but did some simplifications and adjustments for the data source used. The annotated classes with their respective definitions are listed in Table~\ref{tab:class_def} below.

\renewcommand\arraystretch{1.5}
\begin{table}[h]
\centering
\begin{footnotesize}
\begin{tabular}{p{0.4cm}lp{16cm}}
 \toprule
 cat. & class & definition \\
 \midrule
 \multirow{5}{*}{\begin{sideways}flat\end{sideways}}
 & road & Drivable areas where cars could drive on including main road, bike lanes, and crossed areas on the street. Road curb is excluded. \\

 & sidewalk & Paths along sides of the road, used for pedestrians and bicycles, but cars are not allowed to drive on. Also include private driveways.\\

& parking & Areas for parking and are clearly different from sidewalk and road. If unclear then other-ground or sidewalk can be selected. Garages are labeled as building instead of parking.\\

& other-ground & Ground that excludes sidewalk, terrain, road, and parking. It includes (paved/plastered) traffic islands that are not meant for walking. \\

\midrule
\multirow{4}{*}{\begin{sideways}construction\end{sideways}} 
& building & All building parts including walls, doors, windows, stairs, and garages, etc. \\
\\
& fence & Separators including wood or metal fences, small walls and crash barriers. \\
\\
\\
\midrule

\multirow{5}{*}{\begin{sideways}vehicle\end{sideways}} 
& car & Different types of cars, including cars, jeeps, SUVs, and vans.\\

& truck & Trucks, vans with a body that is separate from the driver cabin, pickup trucks, as well as their attached trailers.\\

& bicycle & Including different types of bicycles, without any riders or pedestrians nearby.\\

& motorcycle & Including different types of motorcycles, without any riders or pedestrians nearby. \\

& other-vehicle & Other types of vehicles that do not belong to previously defined vehicle classes, such as various trailers,  excavators, forklifts, and fallbacks.\\
\midrule
\multirow{3}{*}{\begin{sideways}nature\end{sideways}}
 & vegetation & Including bushes, shrubs, foliage, treetop except for trunks, and other clearly identifiable vegetation. \\
 & trunk &  The tree trunk is labeled as \textit{trunk} separately from the treetop.  \\

 & terrain & Mainly include grass and soil.  \\
 
\midrule
\multirow{4}{*}{\begin{sideways}human\end{sideways}} & person & Humans that are standing, walking, sitting, or in any other pose, but not driving any vehicle. Trolley cases, strollers, and pets nearby are excluded.\\
  
 & bicyclist & Humans driving a bicycle or standing in close range to a bicycle (within \textit{arm reach}).
 \\
 
 & motorcyclist & Humans driving a motorcycle or standing in close range to a motorcycle (within \textit{arm reach}). \\
 
\midrule
\multirow{2}{*}{\begin{sideways}object\end{sideways}} 
& pole & Lamp posts, the poles of traffic signs and traffic lights. \\

& traffic sign & Traffic signs excluding their mounting. \\

\midrule

& invalid & Indiscernible semantic contents caused by adverse weather, such as points of thick snow cover, falling snow or rain droplets, and the splash from the rear of the moving vehicles when driving on the road of snow or water.\\

  \bottomrule
  \end{tabular}
 \captionof{table}{Definitions of semantic classes in SemanticSTF.}
 \label{tab:class_def}
\end{footnotesize}
\end{table}

\end{document}